\documentclass[a4paper,fleqn]{cas-dc}

\usepackage[numbers]{natbib}
\usepackage{epstopdf}
\usepackage{color}
\usepackage{bm}
\usepackage{amsmath}
\usepackage{stfloats}  
\usepackage{subfigure}
\usepackage{float}
\usepackage{caption}
\usepackage{multirow}

\begin{document}
    \let\WriteBookmarks\relax
    \def\floatpagepagefraction{1}
    \def\textpagefraction{.001}

\shorttitle{<short title of the paper for running head>}    

\shortauthors{<short author list for running head>}  

\title [mode = title]{An Interactively Reinforced Paradigm for Joint Infrared-Visible Image Fusion and Saliency Object Detection}  

\tnotemark[<tnote number>] 


%











\author[1,4]{\textcolor[RGB]{0,0,1}{Di Wang}}

\author[3]{\textcolor[RGB]{0,0,1}{Jinyuan Liu}}

\author[2,4]{\textcolor[RGB]{0,0,1}{Risheng Liu}}

\author[2,4]{\textcolor[RGB]{0,0,1}{Xin Fan}}

\cormark[1]
\address[1]{School of Software Technology, Dalian University of Technology, Dalian 116620, China.}
\address[2]{DUT-RU International School of Information Science $\&$ Engineering, Dalian University of Technology, Dalian 116620, China}
\address[3]{School of Mechanical Engineering, Dalian University of Technology, Dalian 116023, China}
\address[4]{Key Laboratory for Ubiquitous Network and Service Software of Liaoning Province, Dalian 116620, China}
\ead{xin.fan@dlut.edu.cn}

\cortext[cor1]{Corresponding author: } 













\begin{abstract}
This research focuses on the discovery and localization of hidden objects in the wild and  serves unmanned systems. 
Through empirical analysis, infrared and visible image fusion (IVIF) enables hard-to-find objects apparent, whereas multimodal salient object detection (SOD) accurately delineates the precise spatial location of objects within the picture.  
Their common characteristic of seeking complementary cues from different source images motivates us to explore the collaborative relationship between \textbf{F}usion and \textbf{S}alient object detection tasks on infrared and visible images via an \textbf{I}nteractively \textbf{R}einforced multi-task paradigm for the first time, termed \textbf{IRFS}. 
To the seamless bridge of multimodal image fusion and SOD tasks, we specifically develop a Feature Screening-based Fusion subnetwork (FSFNet) to screen out interfering features from source images, thereby preserving saliency-related features. 
After generating the fused image through FSFNet, it is then fed into the subsequent Fusion-Guided Cross-Complementary SOD subnetwork (FC$^2$Net) as the third modality to drive the precise prediction of the saliency map by leveraging the complementary information derived from the fused image.
In addition, we develop an interactive loop learning strategy to achieve the mutual reinforcement of IVIF and SOD tasks with a shorter training period and fewer network parameters.
Comprehensive experiment results demonstrate that the seamless bridge of IVIF and SOD mutually enhances their performance, and highlights their superiority.
This code is available at~\href{https://github.com/wdhudiekou/IRFS}{https://github.com/wdhudiekou/IRFS}.
\end{abstract}




\maketitle

\section{Inroduction}\label{Intro}

Unmanned system enjoys the merits and presents conduciveness to applications in both military and civilian areas, e.g., unmanned aerial vehicle (UAV), unmanned combat vehicle (UCV), and autonomous driving~\cite{Bogdoll_2022_CVPR}.
However, single-modality vision technologies equipped by some unmanned systems struggle to cope with challenging scenarios in the wild, resulting in the failure to find hidden objects and low-precise localization.
To this end, multimodal sensors are introduced, among which infrared sensors are the most widely-used. Infrared sensors image by thermal radiation emitted from objects and have properties of anti-interference and anti-occlusion, so infrared images can highlight salient objects. In contrast, visible images captured by RGB sensors embrace rich textures and details due to reflective light information. Therefore, infrared and visible image fusion (IVIF) that extracts the complementary information from them to generate a fused image, enables hard-to-find objects apparent and benefits object localization, which is promising to unmanned systems.

\begin{figure}[t]
	\centering
	\begin{tabular}{c}
		\includegraphics[scale=0.33]{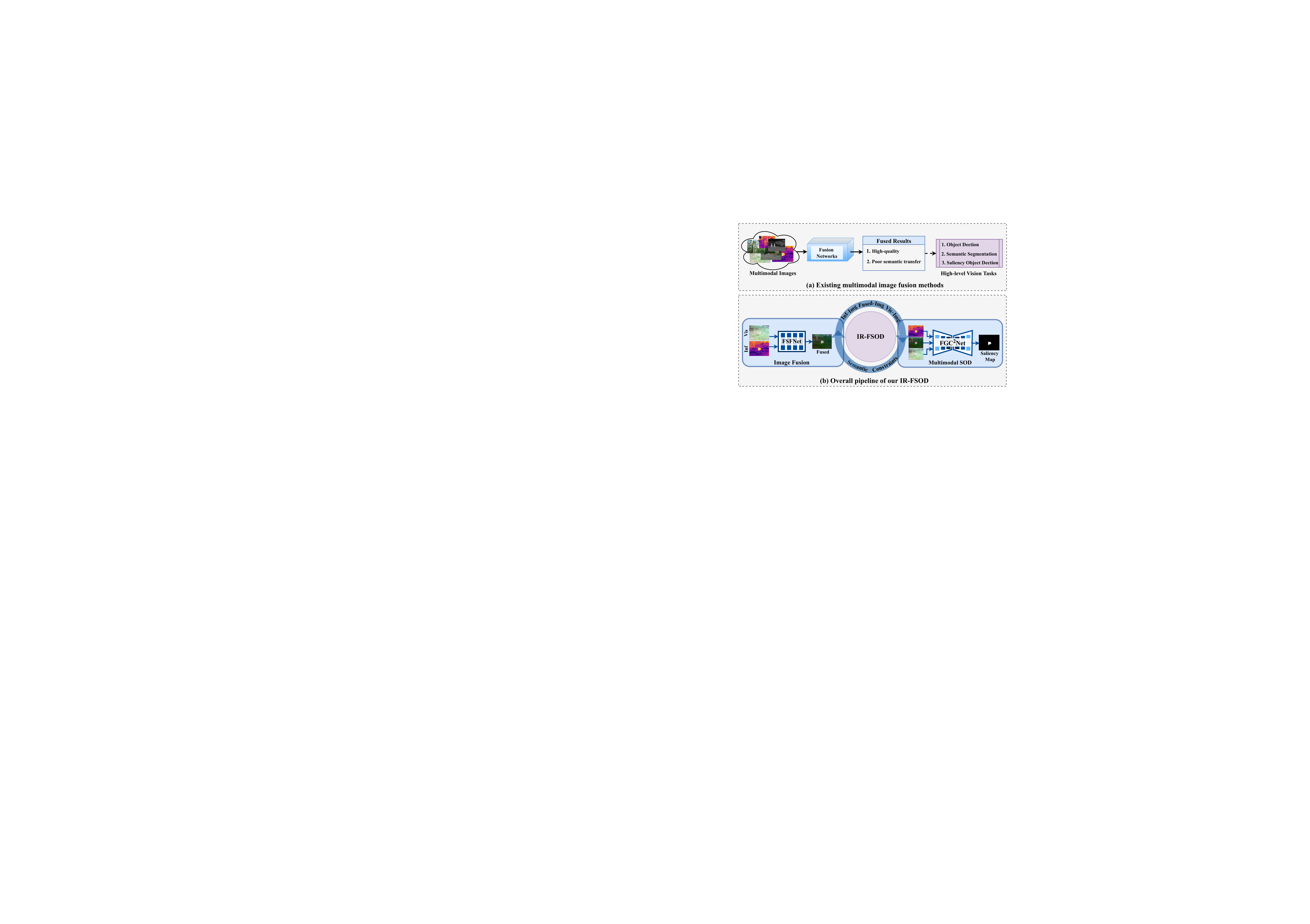} \\
	\end{tabular}
	\vspace{-3mm}
	\caption{Comparison between existing fusion methods and the proposed interactively reinforced paradigms for joint infrared-visible image fusion and saliency object detection (IRFS).}%
	\label{fig:fig_1}
	\vspace{-4mm}
\end{figure}

Prevailing IVIF methods~\cite{FGAN,DDcGAN,GANMcC,PMGI,U2Fusion,STDFusionNet,holoco,coconet,li2023infrared,EMAU} perform extremely well on both highlighting salient objects and manifesting abundant textures. Typically, those fusion methods~\cite{PMGI,DIDFuse_2020,li2023infrared,CDDFuse} based on the proportional preservation of texture and intensity of source images, spatial-guided methods~\cite{STDFusionNet} from annotated salient object masks, and generative adversarial network (GAN)-based methods~\cite{FGAN,DDcGAN,GAN-FM,GANMcC} are included.
However, these methods are one-sided efforts to obtain fusion images with better visual quality, and seldom discuss the adaptability and linkage with downstream high-level vision tasks in practical applications. 
Coupled with existing IVIF methods struggle to transmit precise semantic information into downstream high-level tasks, resulting in a severe performance decline in downstream applications (see Figure~\ref{fig:fig_1}(a)).
Meanwhile, there are some low-level vision methods~\cite{SFTGAN,HDRUNet,jiang2022target,liu2022twin,liu2022learning,ma2021learning} on which high-level semantic information acts. But, they only roughly embed semantic probability maps as a condition into some specific layers, rather than an adaptive bridge between two tasks.

Motivated by the above discussion, we tend to design a joint paradigm to bridge infrared and visible image fusion with a downstream visual task, acting as an adapter.
Considering the common characteristic between infrared-visible salient object detection (SOD) and fusion tasks that seek complementary cues from two source images to predict final results, we make a preliminary attempt to explore the collaborative relationship between them. 
There are three main obstacles to cut through:
(i)~\textit{designing a fusion network that can effectively transmit saliency-related semantic features to cater to saliency object detection.}
Because infrared-visible SOD~\cite{CSRNet_22,MIDD_21,CGFNet_22,ECFFNet_22} aims to extract and fuse hierarchically complementary cues from two source images, thus predicting an accurate binary location map for the most distinctive objects. This relies on semantic information such as the salient structures of objects. 
(ii)~\textit{developing a seamless and efficient bridging manner to push the image fusion to play facilitation in the downstream SOD task}.
Typical multimodal SOD methods often adopt separate feature extractors for each source image and then aggregate the extracted modality-specific features through a complementary fusion module. If we follow this mode, it will inevitably cause a heavyweight model, unexploited modality-shared features, and complex feature aggregation.
(iii)~\textit{devising a collaborative learning strategy and making the two tasks tightly coupled and mutually reinforced.}
Most of the pioneer multi-task methods either follow the low-to-high paradigm or the high-to-low paradigm, often unilaterally ascending one or the other.

In this work, we construct an interactively reinforced paradigm to bridge infrared and visible image fusion and saliency object detection, termed IRFS (see Figure~\ref{fig:fig_1}(b)). Its overall framework consists of a bidirectional optimization stream between image fusion and SOD.
For the low-to-high stream, we specifically design a feature screening-based image fusion network termed FSFNet to screen out interfering features and maintain saliency-related and texture-informative features.
And then, to bridge the two tasks in a more efficient way, the fused image generated by FSFNet is treated as the third modality to guide the downstream SOD. More specifically, we introduce a fusion-guided saliency-enhanced (FGSE) module to perform the reweighting and cross-complementation of source features guided by the fused image. We then embed the FGSE module into each scale of the shared backbone and build a Fusion-guided Cross-Complementary SOD network termed FGC$^{2}$Net, used to keep the guidance effect of the fused image throughout the whole SOD process, thus realizing a seamless bridge between the two tasks. 
For the high-to-low stream, we use the labeled saliency map to supervise the saliency object detector and establish a semantic loss, which is then back-propagated to the fusion sub-network, thereby forcing the generation of fused images with rich semantic information.
In addition, we also develop an interactive loop learning strategy to interactively optimize image fusion and SOD tasks, which finally achieves the optimal results of both with a shorter training period and fewer network parameters.
Our major contributions are concluded as follows:
\begin{itemize}
	\item An interactively reinforced paradigm for joint infrared-visible image fusion and saliency object detection is constructed, in which the collaborative relationship between the two tasks is explored for the first time.
	\item A feature screening-based image fusion network is proposed to highlight the saliency-related semantic features catering to SOD. Meanwhile, a fusion-guided saliency-enhanced module is introduced to transmit the guidance of the upstream fused results throughout the downstream SOD task. Thus, the seamless bridge between the two tasks is achieved.
	\item We devise an interactive loop learning strategy to tightly couple the fusion and SOD tasks on infrared-visible images, attaching an optimal balance of them with the least training costs.
\end{itemize}
Experiment results show that the proposed IRFS can seamlessly bridge infrared-visible image fusion and SOD tasks, and both of them benefit from each other and perform superior capabilities.

\section{Related Works}\label{Rew}
Multimodal image fusion and multimodal salient object detection are the supporting technologies of this work. In this section, We survey adequately existing technologies and review their development.

\subsection{Multimodal Image Fusion}
%
%
In the past few decades, multimodal image fusion has made significant progress, which roughly falls into two categories, i.e., traditional and deep learning-based methods.

\subsubsection{Traditional image fusion methods}
Typical traditional methods contain multi-scale transform (MST)-based~\cite{MST_22}, sparse representation-based~\cite{SRFusion}, subspace-based~\cite{CvejicLBC06}, and saliency-based~\cite{saliency-based} fusion methods.

MST-based image fusion methods comprise three steps, i.e., the decomposition of multi-scale features, the fusion of multi-scale features, and the inversion of fused features. 
Chen et al.~\cite{MST_22} followed the pipeline and leveraged laplacian pyramid transformation to conduct multi-scale feature decomposition. In addition to the laplacian pyramid, wavelet transform~\cite{wavelet}, contourlet transform~\cite{nsct}, and edge-preserving filters~\cite{edge_pre} are also widely used in MST-based image fusion.

Sparse representation-based image fusion methods~\cite{sparse_1, sparse_2} usually rely on an over-complete dictionary learned from numerous high-quality natural images. 
Concretely, sparse representations are encoded from each source image through the learned over-complete dictionary, and then the encoded fused sparse coefficients are transformed into a fused image via this over-complete dictionary. Moreover, sparse coding patterns are also various.
For instance, Bin et al.~\cite{BinCG16} proposed an approximate sparse representation using a multi-selection strategy to obtain sparse coefficients from source images. Liu et al.~\cite{LiuCWW16} utilized convolutional sparse representation to complete the acquisition of sparse coefficients.

Subspace-based image fusion methods are to capture the native structures of the source image by mapping higher-dimensional inputs into a lower-dimensional space or subspace, which involves different dimensionality reduction manners, e.g., Principal Component Analysis (PCA), Independent Component Analysis (ICA), and Non-negative Matrix Factorization (NMF).
For instance, Li et al.~\cite{LI201628} leveraged PCA to fuse decomposed low-frequency images, while Bavirisetti et al.~\cite{BavirisettiXL17} adopted PCA to fuse high-frequency detail images.
Cvejic et al.~\cite{CvejicLBC06} proposed to segment source images into different regions and then obtain ICA coefficients for each region through a set of bases pretrained from natural images.
Mou et al.~\cite{NMF} proposed to extract features from infrared and visible images by NMF, which is capable of preserving textures of visible images and high-contrast structures of infrared images while removing noise.

Saliency-based image fusion methods are inspired by human visual attention, and they are conducive to highlighting salient objects in fused images.
There are two types of saliency-based fusion methods, i.e., salient object extraction and weight calculation.
The first type of methods, extracting the salient regions from source images and mapping them to the fused images, are capable of preserving dominant information, such as~\cite{LIU201794} and~\cite{QU20081508}.
The other type of methods need to first obtain salient weight maps for the base and detail image layers, respectively, and then obtain base and detail images through the weighted combination between image layers with their weight maps, such as~\cite{GAN201537}.

Although visually favorable fused images are generated, these traditional image fusion methods are still inferior to deep learning-based methods.

\subsubsection{Deep-learning based image fusion methods}
Deep learning-based methods are divided into four categories: autoencoder (AE)-based~\cite{DenseFuse}, deep CNN-based~\cite{PMGI, holoco}, generative adversarial network (GAN)-based~\cite{FGAN, DDcGAN, GANMcC}, and transformer-based image fusion methods.

AE-based image fusion methods follow a common Encoder-Decoder paradigm to accomplish the extraction of multimodal features and the reconstruction of the fused image. As a pioneer, Li et al.~\cite{DenseFuse} proposed dense blocks as the fundamental components of the auto-encoder to perform feature extraction, then fuse extracted multimodal features via rough addition fusion and $l_{1}$-norm fusion rules. The fused features are reconstructed as a fused image by a plain decoder. 
Since successive downsampling operations in auto-encoders leads to loss of effective information, Li et al.~\cite{RFN} introduced residual connections into the Encoder-Decoder paradigm to alleviate this problem.
To explore the interpretability of the auto-encoder, Zhao et al.~\cite{DIDFuse_2020} proposed to disentangle background and detail features by implementing two-scale decomposition and extracting low- and high-frequency features in an encoder.

Deep CNN-based image fusion methods mainly focus on the design of a variety of network architectures and fusion strategies.
Zhang et al.~\cite{PMGI} introduced a dual-path dense network to learn intensity and gradient features from multimodal images separately, and then devised specific losses to maintain the balance of intensity and gradient information in fused images.
Still following dense network structures, Xu et al.~\cite{U2Fusion} combined a dense network and an information measurement in a unified framework, which is capable of estimating the importance of different image sources adaptively and can be used to solve various fusion problems.
However, simple dense networks do not perform well in preserving high-quality structures and details.
Subsequently, Liu et al.~\cite{MFEIF} applied a coarse-to-fine network structure to extract multi-scale features, which has a stronger feature representation ability than the plain dense network. Liu et al. also designed an edge-guided attention module to push the network to highlight prominent structures and preserve abundant details better.
The above manual networks exhibit a lack of flexibility when faced with different types of image data. To this end, Liu et al.~\cite{LiuLL021} first utilized Neural Architecture Search (NAS) methodology to build a hierarchically aggregated fusion architecture, which is more flexible and effective for different fusion demands. Some relevant follow-up studies, e.g.,~\cite{SMoA_21} and~\cite{LiuWWLF22} applied NAS to search a Modality-orient network and a lightweight target-aware network for infrared and visible image fusion. 

GAN-based image fusion methods aim to build constraints from the perspective of a probability distribution, so as to achieve sharp targets and rich textures of fused images while reaching a balance of information transmission from source images.
Ma et al.~\cite{FGAN} first introduced the generative adversarial network into the field of image fusion, modeling this task in an adversarial manner between a generator and a discriminator. However, such single-discriminator models~\cite{FGAN, GANMcC} force the generator to equally treat different modalities, resulting in over-smoothed fused images.
Thus, several dual-discriminator image fusion networks are proposed, such as~\cite{DDcGAN, GAN-FM, AGAL}, which are capable of highlighting high-contrast structures and fine-grained textures.

Transformer-based image fusion methods~\cite{PPTFusion, SwinFusion, CGTF} have emerged in the past two years since the transformer embraces a global receptive field and is able to model long-range dependencies of neighboring pixels.
Due to the fact that existing transformers neglect the local spatial correlation between pixels, Fu et al.~\cite{PPTFusion} first proposed a Patch Pyramid Transformer (PPT) framework, in which the patch transformer is used to model local feature representations and pyramid transformer is used to model non-local feature representations.
Subsequently, Li et al.~\cite{CGTF} proposed a convolution-guided transformer aimed to first leverage a convolutional feature extractor to learn local features and then use them to guide the transformer-based feature extractor to capture long-range interdependencies of features.  
With the prevalence of Swin Transformer~\cite{SwinTrans}, Ma et al.~\cite{SwinFusion} designed a unified multi-task fusion framework based on the shifted windows mechanism from~\cite{SwinTrans} and self- and cross-attention mechanisms.
Such methods perform well in preserving image structures and details.

Unfortunately, the aforementioned methods pay one-sided attention to the visual quality while failing to establish a connection with high-level vision tasks. 
Considering the comprehensive understanding ability of the unmanned system in the wild, it is imperative to exploit a joint framework for infrared-visible image fusion collaborated with the high-level vision task. 

\subsection{Multimodal Salient Object Detection}
Recent years have witnessed great progress in thermal infrared and visible saliency object detection (SOD) with the popularity of thermal infrared sensors in the field of multimodal SOD, including traditional~\cite{M3S-NIR_19, SGDL_20, MTMR_18} and deep learning-based methods~\cite{ADF_20, CSRN_21, MIDD_21, CGFNet_22}.
Traditional thermal infrared and visible SOD methods comprise ranking-based and graph learning-based methods. 
Wang et al.~\cite{MTMR_18} first applied a ranking algorithm to the multimodal SOD task and proposed a  multi-task manifold ranking pattern. And, they built a thermal infrared and visible benchmark termed VT821.
Graph learning-based SOD methods, representatively, Tu et al.~\cite{SGDL_20} proposed a collaborative graph learning model, in which source images are segmented into superpixels as graph nodes and then learn graph affinity and node saliency.
Subsequently, Tu et al.~\cite{M3S-NIR_19} also proposed to conduct a graph-based manifold ranking model in a set of multi-scale superpixels by combining graph learning and ranking theory, and optimizing the model based on ADMM optimizer.

Owning to the representation ability of CNNs, a basic idea of deep learning-based thermal infrared-visible SOD methods is to extract complementary information from source images to predict accurate saliency maps of conspicuous objects.
Tu et al.~\cite{ADF_20} built a large-scale thermal infrared-visible benchmark termed VT5000, and proposed a dual-encoder framework to extract complementary features from different modalities. 
To better explore interactions between multimodal features, Tu et al.~\cite{MIDD_21} proposed a dual-decoder framework to perform interactions of multi-scale features, which is more favorable for challenging scenarios.
Seeing that such dual-encoder and dual-decoder methods have a larger model size, Liao et al.~\cite{CCEFN} devised a single-encoder paradigm to use a shared encoder to extract complementary features from thermal infrared and visible images. 

However, either traditional or deep infrared-visible SOD methods only strive for cross-modal interaction and fusion on feature space, but never pixel-level fusion. 
In practice, the fused image can highlight structures of objects, which also play a critical role in distinguishing salient objects.
Therefore, it is natural to consider the combination of image fusion and SOD tasks in a single framework to achieve mutual benefit.

\subsection{Joint Frameworks of Multiple Vision Tasks}
Recently, some practical demands have promoted the incorporation of low- and high-level vision tasks. 
One route is to establish a low-to-high cascaded pipeline to allow the low-level task to facilitate the high-level task~\cite{derain_seg_20,denoise_high_20,TangYLT22,ZhaTST23,LiTPQT23,TangLPT20}.
Another route is embedding semantic probability maps as a condition into some specific layers to provide high-to-low guidance for some low-level restoration tasks, e.g., image super-resolution~\cite{SFTGAN,TCGD}, image enhancement~\cite{jiang2022towards,liu2021retinex,ma2022toward}, and image HDR~\cite{HDRUNet}.
A third route is to build a parallel framework to equally treat the low- and high-level vision tasks~\cite{DSNet_21, Dual_super_seg_20}.
Unfortunately, none discusses the adaptive bridge between tasks, resulting in overfitting one of the tasks and deviating from the optimal balance between them.
Two recent studies~\cite{SeAFusion, TarDAL_22}, in fact, explored the relationship between image fusion and high-level tasks. However, \cite{TarDAL_22} only considers the high-level task-oriented joint training, and~\cite{SeAFusion} only focuses on the trade-off between low- and high-level losses.
The latest study, termed SuperFusion~\cite{SuperFusion}, integrated image registration, image fusion, and semantic segmentation into a unified framework.
In the framework, image registration and fusion are jointly optimized in a symmetric scheme, enabling the two tasks to mutually promote each other. A pretrained semantic segmentation model then was deployed to guide the fusion network to focus more on semantic-related features.
However, these methods ignore the intrinsic relation between pixel-level fused results and multimodal high-level vision tasks, which becomes the focus of this work.


\section{The Proposed Method}
The overview of the proposed interactively reinforced paradigm for joint infrared and visible image fusion and saliency object detection is shown in Figure~\ref{fig:pipeline}.
This paradigm contains two sub-tasks, where image fusion is regarded as the dominant task, while the multimodal saliency object detection task is treated as a downstream task of image fusion and facilitates saliency-oriented image fusion as an auxiliary tool.
The overall network structure contains a feature screening-based image fusion subnetwork (FSFNet) and a fusion-guided cross-complementary SOD subnetwork (FGC$^{2}$Net).

\subsection{Feature Screening-based Image Fusion}
\begin{figure*}[tbh]
	\centering
	\begin{tabular}{c}
		\includegraphics[width = 0.90\linewidth]{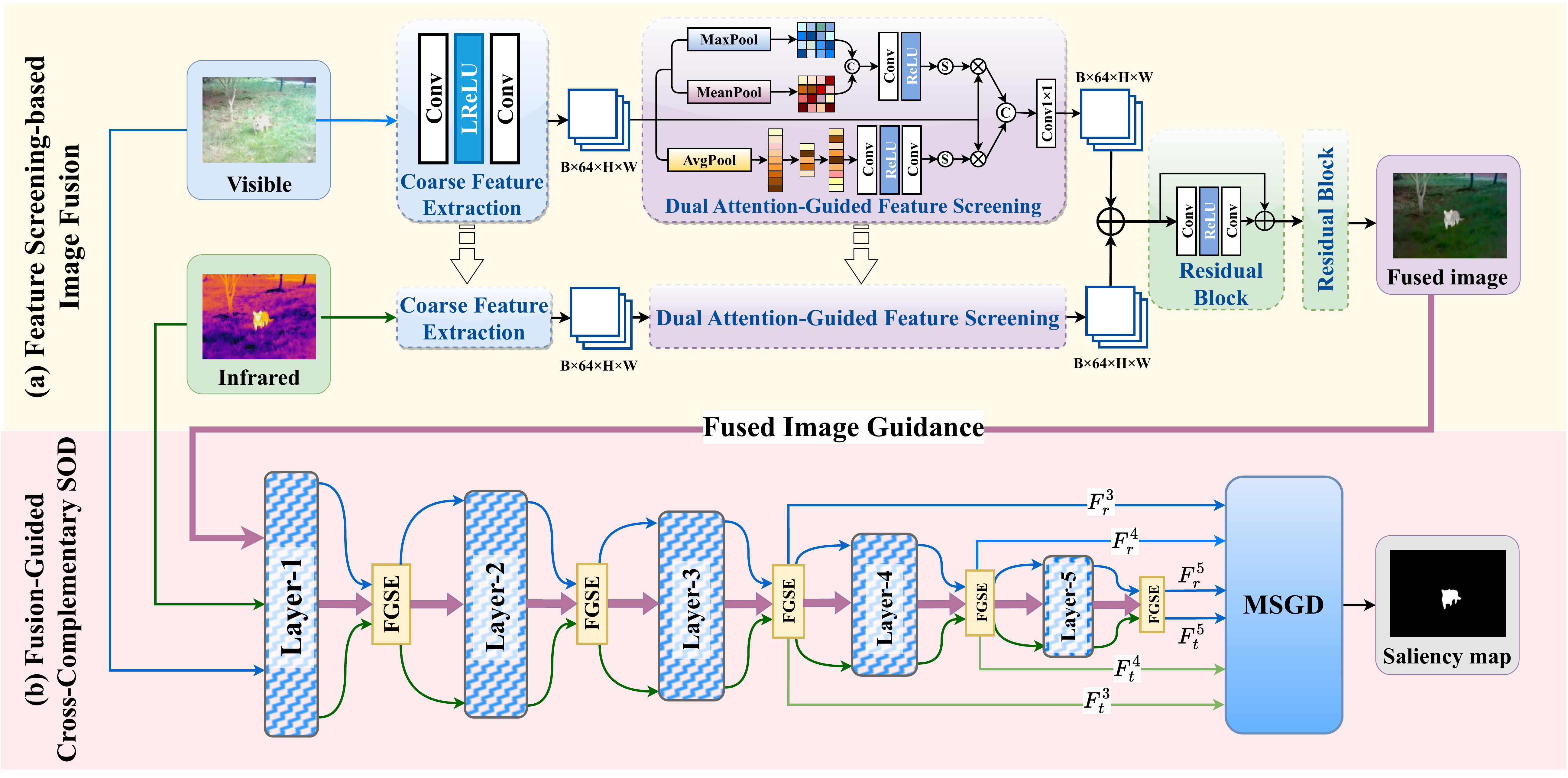} \\
	\end{tabular}
	\vspace{-2mm}
	\caption{Overview of the proposed interactively reinforced paradigm for joint infrared-visible image fusion and saliency object detection called IRFS. The paradigm is a cascaded framework that consists of a feature screening-based image fusion subnetwork termed FSFNet (as shown in (a)) and a fusion-guided cross-complementary SOD subnetwork termed FGC$^{2}$Net (as shown in (b)). In this framework, the image fusion facilitates the SOD task from the bottom up, and conversely, SOD facilitates the image fusion task from the top down.}%
	\label{fig:pipeline}
	\vspace{-4mm}
\end{figure*}

Aiming at the goal that fused images promote salient object detection in our IRFS framework, a specific fusion network is designed, as shown in Figure~\ref{fig:pipeline}(a), which can not only generate high-quality fused images, but also keep the semantic information in fused images.
Given a pair of visible image $I_{r}\in \mathbb{R}^{H \times W \times 3}$ and infrared image $I_{t}\in \mathbb{R}^{H \times W \times 1}$, we first utilize a coarse feature extractor $\mathcal{F}^{c}\left(\cdot\right)$ consisting of two convolution layers and a Leaky ReLU activation function to extract coarse features $\textbf{F}_{c}=\left[\textbf{F}_{c}^{r}, \textbf{F}_{c}^{t}\right]$.
Noted that the visible image needs to be converted to YCbCr color space, and then taking the Y-channel image as the input of the visible branch. 
Next, it is necessary to study how to screen out interference features and preserve saliency-related features from $\textbf{F}_{c}$ to facilitate the subsequent SOD task. 
Due to attention mechanisms~\cite{Hu_2018_SENet_CVPR} can model the feature correlations in both channel and spatial dimensions, and are conducive to capturing fine-grained texture features, we are determined to deploy a dual attention-guided feature screening module (DAFS) to screen out useless features and preserve saliency-related and texture-informative precise features, thus catering to the requirement of SOD task. Therefore, these precise features can be formulated as
\begin{equation}
	\begin{split}
		\textbf{F}_{\text{p}}^{r}&=Conv_{1\times 1}\left(\left[SA\left(\textbf{F}_{c}^{r}\right), CA\left(\textbf{F}_{c}^{r}\right)\right]\right),\\
		\textbf{F}_{\text{p}}^{t}&=Conv_{1\times 1}\left(\left[SA\left(\textbf{F}_{c}^{t}\right), CA\left(\textbf{F}_{c}^{t}\right)\right]\right).
	\end{split}
\end{equation}
Then, we fuse preserved features from source images by
\begin{equation}
	\begin{split}
		\textbf{F}_{\text{u}}^{r}= \textbf{F}_{\text{p}}^{r} \oplus \textbf{F}_{\text{p}}^{t},
	\end{split}
\end{equation}
where $\oplus$ denotes element-wise summation operation.
We adopt serial residual blocks to reconstruct the fused Y-channel image as $I_{f}^{y}$. We then convert $I_{f}^{y}$ to the RGB image $I_{f} \in \mathbb{R}^{H \times W \times 3}$ so that it serves the subsequent SOD task.

\subsection{Fusion-Guided Cross-Complementary SOD}

Benefiting from the fused images with sharp objects and the high contrast between objects and surroundings, we treat the fused image as the third type of modality to guide the infrared-visible SOD task. This is the first attempt at breaking out of the standard multi-modality SOD configurations.
As shown in Figure~\ref{fig:pipeline}(b), we propose a fusion-guided cross-complementary SOD network, termed as FGC$^{2}$Net, which takes a group of images $\left\{I_{r}, I_{f}, I_{t}\right\}$ as input and is supposed to predict a precise saliency map $\textbf{M}_{p}$ for the most conspicuous objects.

\begin{figure}[t]
	\centering
	\begin{tabular}{c}
		\includegraphics[width = 0.99\linewidth]{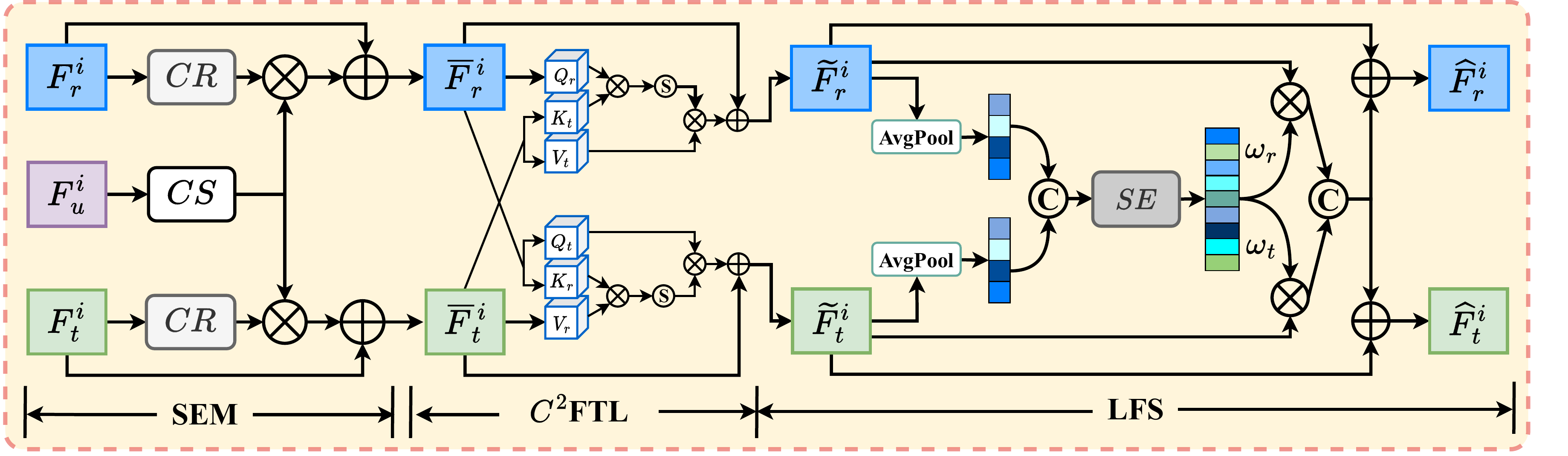}\\
	\end{tabular}
	\vspace{-2mm}
	\caption{The details of the proposed FGSE module. This module consists of three components including a saliency-enhanced module (SEM), a cross-complementary feature transformer layer (C$^2$FTL), and a learnable feature selector (LFS).}%
	\label{fig:FGSE}
	\vspace{-4mm}
\end{figure}

Different from some impressive works~\cite{MIDD_21,MMNet_21} that assign an individual backbone for infrared and visible images, respectively, to extract cross-modal hierarchical features and then fuse them step-by-step in an extra branch, our FGC$^{2}$Net employs a siamese encoder to alternately perform feature extraction and cross-modality feature aggregation. 
In particular, we introduce a fusion-guided saliency-enhanced (FGSE) module to conduct cross-modality feature aggregation and embed it behind each feature scale of the backbone. The purpose is to reweight the infrared and visible images using the high contrast differences of objects and backgrounds of the fused image, further enhancing saliency-related features and suppressing surrounding interference features.
As shown in Figure~\ref{fig:FGSE}, the FGSE is divided into three steps. Given the feature set $\{F_{r}^{i-1}, F_{t}^{i-1}, F_{u}^{i-1}\}$ from the $\left(i-1\right)$-th scale of the backbone, saliency-enhanced infrared and visible image features are first obtained under the guidance of the fused feature $f_{u}^{i}$ and formulated as
\begin{equation}
	\begin{split}
		\overline{F}_{r}^{i}&=F_{r}^{i-1} \oplus \mathcal{CR}\left(F_{r}^{i}\right)\otimes \mathcal{CS}\left(F_{u}^{i}\right),\\
		\overline{F}_{t}^{i}&=F_{t}^{i-1} \oplus \mathcal{CR}\left(F_{t}^{i}\right)\otimes \mathcal{CS}\left(F_{u}^{i}\right),
	\end{split}
\end{equation}
where, $\mathcal{CR}\left(\cdot\right)$ denotes convolution and ReLU function, and $\mathcal{CS}\left(\cdot\right)$ denotes convolution and sigmoid function. The $\oplus$ and $\otimes$ represent element-wise summation and multiplication, respectively.
Based on the self-attention mechanism, we then introduce a cross-complementary feature transformer layer (C$^2$FTL) to learn cross-modality structure-sharp and texture-informative features from $\overline{F}_{r}^{i}$ and $\overline{F}_{t}^{i}$. 
Before that, $\overline{F}_{r}^{i}$ is transformed into $\textbf{Q}_{r}$, $\textbf{K}_{r}$, and $\textbf{V}_{r}$, while doing the same for $\overline{F}_{r}^{i}$. Then, we adopt the C$^2$FTL to generate complementary cross-modal features by
\begin{equation}
	\begin{split}
		\widetilde{F}_{r}^{i}&=\mathcal{S}\left(\textbf{Q}_{r} \otimes \textbf{K}_{t}\right)\otimes \textbf{V}_{t} \oplus F_{r}^{i-1},\\
		\widetilde{F}_{t}^{i}&=\mathcal{S}\left(\textbf{Q}_{t} \otimes \textbf{K}_{r}\right)\otimes \textbf{V}_{r} \oplus F_{t}^{i-1},
	\end{split}
\end{equation}
where, $\mathcal{S}\left(\cdot\right)$ is sigmoid function.

To prevent further forward propagation of interfering information in $\widetilde{F}_{r}^{i}$ and $\widetilde{F}_{t}^{i}$, we deploy a learnable feature selector (LFS) to suppress saliency-irrelevant features. 
The core of the feature selection is to generate a weight vector based on the global average pooling operation and the feature squeeze-and-excitation (SE) operation, and then generate two learnable parameters by the softmax function.
%
%
This process can be formulated as
\begin{equation}
	\begin{split}
		\omega_{r}, \omega_{t}=Softmax\left(SE\left[\mathcal{P}(\widetilde{F}_{r}^{i}), \mathcal{P}(\widetilde{F}_{t}^{i})\right]\right),
	\end{split}
\end{equation}
where, $\mathcal{P}\left(\cdot\right)$ refers to the global average pooling and $\left[\cdot\right]$ refers to the concatenation along with channel dimension.
Immediately, we utilize $\omega_{r}$ and $\omega_{t}$ to reweight $\widetilde{F}_{r}^{i}$ and $\widetilde{F}_{t}^{i}$ to reconstruct saliency-enhanced cross-modality feature as
\begin{equation}
	\begin{split}
		F_{r}^{i}&=\left[\omega_{r} \otimes \widetilde{F}_{r}^{i}, \omega_{t} \otimes \widetilde{F}_{t}^{i}\right] \oplus \widetilde{F}_{r}^{i},\\
		F_{t}^{i}&=\left[\omega_{r} \otimes \widetilde{F}_{r}^{i}, \omega_{t} \otimes \widetilde{F}_{t}^{i}\right] \oplus \widetilde{F}_{t}^{i}.
	\end{split}
\end{equation}
Noted that, after the concatenation, a followed $1\times1$ convolution to reduce the dimension of features to match them with the inputs (i.e., $\widetilde{F}_{r}^{i}$ and $\widetilde{F}_{t}^{i}$).

\begin{figure}[t]
	\centering
	\begin{tabular}{c}
		\includegraphics[width = 0.95\linewidth]{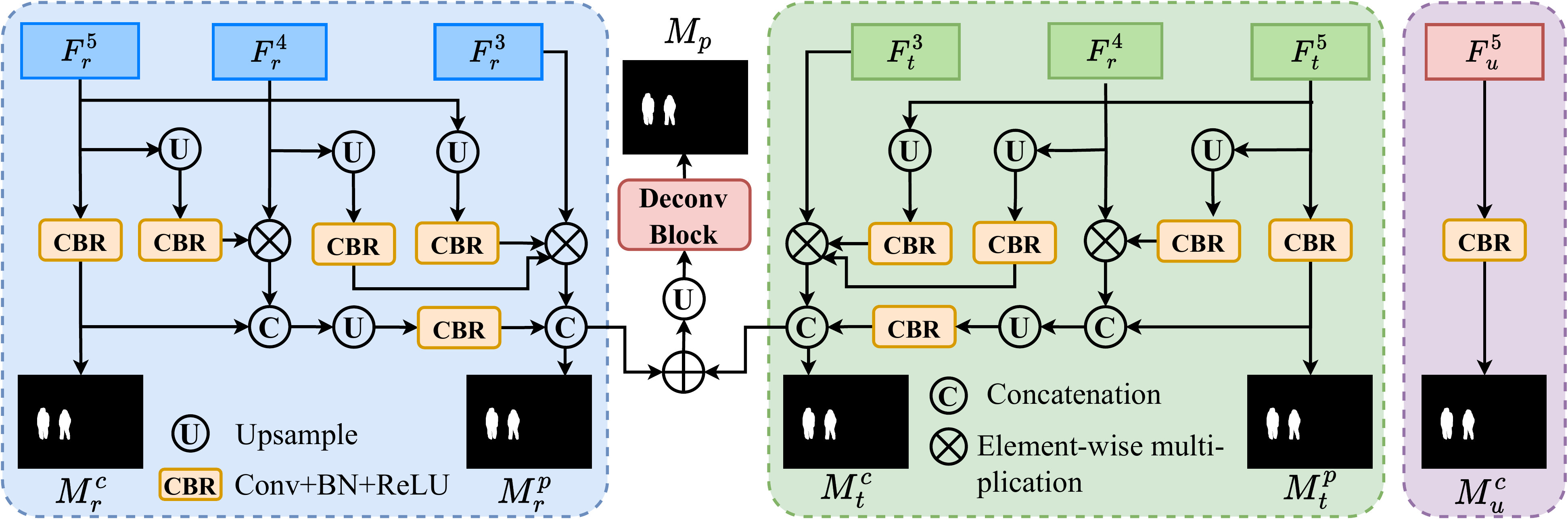} \\
	\end{tabular}
	\caption{Schematic diagram of the modality-specific group decoder (MSGD).}%
	\label{fig:sod_decoder}
	\vspace{-4mm}
\end{figure}

Through the siamese encoder, a hierarchical feature set $\{F_{r}^{i}, F_{t}^{i}, F_{u}^{i} |i \in \{1,2, \ldots, 5\}\}$ is learned.
Next, we introduce a modality-specific group decoder (MSGD) to predict saliency maps, as shown in Figure~\ref{fig:sod_decoder}.
To reduce the computational burden, we only input the features $\{F_{r}^{i}, F_{t}^{i}, F_{u}^{i} |i \in \{3,4,5\}\}$ into MSGD. 
The group decoder consists of three modality-specific decoding branches, i.e., infrared-, visible-, and fusion-modality decoders. Where the fusion-modality decoding branch only predicts a coarse saliency map $M_{u}^{c}$ by a Conv+BN+ReLU (CBR) layer, while infrared- and visible-modality decoding branches simultaneously predict both coarse maps (e.g., $M_{r}^{c}$, $M_{t}^{c}$) and precise maps (e.g., $M_{r}^{p}$, $M_{t}^{p}$).
%
%
Finally, the precise modality-specific saliency maps (i.e., $M_{r}^{p}$, $M_{t}^{p}$) are aggregated to generate the final precise saliency map $M_{p}$.
Noted that the all of features to be decoded go through a global context module and cascaded CBR layers to generate saliency maps in the MSGD module.

\subsection{Loss Functions}
\textbf{Fusion Losses.}
In the fusion phase, we improve the visual quality of the fused results from intensity and gradient perspectives, and the fusion loss is defined as
\begin{equation}
	\begin{split}
        \label{equ:fus_loss}
        \mathcal{L}_{\texttt{fusion}}=\mathcal{L}_{\texttt{int}}+\lambda\mathcal{L}_{\texttt{grad}},
	\end{split}
\end{equation}
where $\lambda$ is a trade-off parameter.
To retain the saliency objects from RGB and thermal images, we build saliency-related intensity loss inspired by~\cite{SMoA_21,TarDAL_22}. The intensity loss is defined as
\begin{equation}
\label{equ:int_loss}
	\begin{split}
		\mathcal{L}_{\texttt{int}}=\|\left(\boldsymbol{\omega}_{t} \otimes I_{t}+\boldsymbol{\omega}_{r} \otimes I_{r}\right), I_{f} \|_{1}+\\
		\gamma\left(1-\texttt{SSIM}\left(\left(\boldsymbol{\omega}_{t} \otimes I_{t}+\boldsymbol{\omega}_{r} \otimes I_{r}\right), I_{f}\right)\right).
	\end{split}
\end{equation}
It relies on $L_{1}$ norm and MS-SSIM to measure the pixel-level similarity between the fused image and source images. Here, $\boldsymbol{\omega}_{r}$ and $\boldsymbol{\omega}_{t}$ are weighted maps of RGB and thermal images, which are obtained through $\boldsymbol{\omega}_{t}=\boldsymbol{S}_{f_{t}} /\left(\boldsymbol{S}_{f_{t}}-\boldsymbol{S}_{f_{r}}\right)$, and $\boldsymbol{\omega}_{r}=1-\boldsymbol{\omega}_{t}$, respectively. $\boldsymbol{S}$ stands for the saliency matrix computed by~\cite{SvSW}.

To preserve fine-grained textures of the fused images, we impose a constraint on the gradient distribution of the fused image and source images as 
\begin{equation}
	\begin{split}
		\mathcal{L}_{\texttt{grad}}=\| \nabla I_{f}, max\left(\nabla I_{r}, \nabla I_{t}\right)\|_{1},
	\end{split}
\end{equation}
where $\nabla$ refers to the Laplacian gradient operator. $max(\cdot)$ denotes the maximum aggregation of fine-grained textures of two source images.

\textbf{SOD Losses.} In the SOD phase, we use the weighted binary cross entropy (wBCE) loss and weighted IoU (wIoU) to supervise the FGC$^2$Net. Given coarse saliency maps $\textbf{M}_{c}=\left[M_{r}^{c}, M_{t}^{c}, M_{u}^{c}\right]$ from the top layer of the siamese encoder, the coarse loss is computed by 
\begin{equation} 
	\mathcal{L}_{\texttt{coarse}}=\mathcal{L}_{bce}^{w}\left(\textbf{M}_{c},GT\right) + \mathcal{L}_{iou}^{w}\left(\textbf{M}_{c},GT\right).
\end{equation}
Then, given precise saliency maps $\textbf{M}_{p}=\left[M_{r}^{p}, M_{t}^{p}\right]$ from the end of the MSGD, the precise loss is computed by
\begin{equation} 
	\mathcal{L}_{\texttt{precise}}=\mathcal{L}_{bce}^{w}\left(\textbf{M}_{p},GT\right) + \mathcal{L}_{iou}^{w}\left(\textbf{M}_{p},GT\right).
\end{equation}
The SOD loss can be formulated as
\begin{equation} 
	\label{equ:sod_loss}
	\mathcal{L}_{\texttt{sod}}=\mathcal{L}_{\texttt{coarse}}+\mathcal{L}_{\texttt{precise}}.
\end{equation}
Therefore, the overall loss is defined as
\begin{equation} 
	\label{equ:joint_loss}
	\mathcal{L}_{\texttt{overall}}=\tau\mathcal{L}_{\texttt{fusion}}+\eta\mathcal{L}_{\texttt{sod}},
\end{equation}
where $\tau$ and $\eta$ are trade-off weights. $\eta$ is initially set to $1$ and increases with the loops of interactive learning, while $\tau$ is identically set to $1$.
We discuss the influence of different values of $\tau$ in Subsection~\ref{subsec:tau}.

\subsection{Interactive Loop Learning Strategy}
In the multi-task frameworks, in general, they are accustomed to adopting a one-stage training manner to find optimal results.
For instance, taking the multi-modal images into the fusion network, the generated fused image is then passed through the SOD network, and the joint loss is calculated to update the two parts of the framework at the same time.
However, this training manner often struggles to achieve a balance between tasks.
To solve this problem, we devise an interactive loop learning strategy. Specifically, when optimizing the fusion part, in order to reinforce the semantic information of the fused image and thus cater to the subsequent SOD task, we force the fused image to pass FSFNet and use Eq.~(\ref{equ:joint_loss}) to update the parameters of the FSFNet. In this case, the SOD network is frozen without updating parameters.
Interactively, when optimizing the SOD part, we use the generated fused image to guide the extraction of saliency-related multimodal features from source images, and the parameters of FGC$^2$Net are updated under the constraint of Eq.~(\ref{equ:sod_loss}). In the meantime, the gradients are truncated at the end of the fusion part to prevent gradient back-propagation from interfering with the optimization of the fusion part.
The interactive loop training process is performed $m$ times in total.
In each loop, the fusion network goes through $n_{f}$ epochs and the SOD network goes through $n_{s}$ epochs.
In this way, the performance of image fusion and SOD tasks can reach an optimal balance in the shortest possible training period.

\section{Experiments}
This chapter first provides detailed descriptions of implementation details, datasets, and evaluation metrics.
Then, the quantitative and qualitative evaluations of our IRFS in the joint multimodal image fusion and SOD task are performed.
Additionally, we evaluate the generalization ability of the IRFS in the aforementioned two sub-tasks.
Lastly, we undertake ablation studies on each key component of our IRFS.

\subsection{Experimental Setup}
\subsubsection{Datasets} 
To validate the effectiveness of the proposed IRFS framework, we jointly evaluate fusion and SOD results on VT5000~\cite{ADF_20} dataset. 
The VT5000 dataset collects 5,000 pairs of thermal infrared and visible images, as well as corresponding binary labels dedicated to SOD, of which 2,500 pairs are used for training and the other 2,500 pairs are used for testing. Therefore, it is convincing that apply the VT5000 dataset to the evaluation of joint multimodal image fusion and SOD tasks.
In addition, we also intend to separately validate the generalization of the proposed IRFS on thermal infrared and visible image fusion and SOD tasks.
In the image fusion task, we use three public fusion datasets including TNO\footnote{http://figshare.com/articles/TNO\_Image\_Fusion\_Dataset/1008029.}, RoadScene\footnote{https://github.com/hanna-xu/RoadScene.}, and M$^3$FD~\cite{TarDAL_22} to directly generate fused images by the pretrained fusion sub-network (i.e., FSFNet) of our IRFS framework, instead of finetuning it with the aforementioned datasets.
Similarly, in the SOD task, we use two benchmarks (i.e., VT1000~\cite{SGDL_20} and VT821~\cite{MTMR_18}) in addition to VT5000 to directly predict final saliency maps taking fused images generated by FSFNet as the third input.
%
%
%

\subsubsection{Implement Details} 
The proposed IRFS scheme is implemented on an NVIDIA 1080Ti GPU using PyTorch framework, in which FSFNet and FGC$^{2}$Net are interactively trained. 
In the training phase, we randomly select $8$ images and resize them to $352\times352$ to form an input batch. Random horizontal flipping is used to reduce over-fitting. 
The parameters in the overall IRFS are updated with the Adam optimizer. In each loop of interactive training for FSFNet and FGC$^{2}$Net, the learning rate of the former is initially set to $1e-3$ and keeps unchanged, while that of the latter is initially set to $5e-5$ and gradually decreases to $1e-6$ following the cosine annealing strategy. In addition, the siamese encoder of FGC$^{2}$Net is dependent on the pretrained ResNet-34~\cite{ResNet34} backbone.
Regarding the setting of hyper-parameters, $\lambda$ in Eq.~(\ref{equ:fus_loss}) is set to $0.5$, and $\gamma$ in Eq.~(\ref{equ:int_loss}) is set to $20.0$.
During the interactive loop learning, $m$ is set to $10$, and $n_{f}$ and $n_{s}$ are set to $3$ and $10$, respectively. Accordingly, $\eta$ in Eq.~(\ref{equ:joint_loss}) increases from $1$ to $10$.

\subsubsection{Evaluation metrics} \label{sec:metric}
We adopt three common metrics to evaluate the quality of the fused image including $\text{MI}$~\cite{MI}, $\text{VIF}$~\cite{vif}, and $\text{CC}$.
%

\textbf{MI.} Mutual Information (MI), derived from information theory, measures the amount of information transmitted from input images to the final fused image, which is calculated by
\begin{equation}
	\begin{split}
		\text{MI}=\text{MI}_{I_{ir}, I_{f}} + \text{MI}_{I_{vis}, I_{f}},
	\end{split}
\end{equation}
where $\text{MI}_{I_{ir}, I_{f}}$ and $\text{MI}_{I_{vis}, I_{f}}$ denote the amount of information that is transmitted from source infrared image $I_{ir}$ and visible image $I_{vis}$ to the fused image $I_{f}$, respectively. $\text{MI}_{I_{ir/vis}, I_{f}}$ is calculated by the Kullback-Leibler measure as follows:
\begin{equation}
	\begin{split}
		\text{MI}_{I_{ir/vis}, I_{f}}=\sum p(I_{ir/vis}, I_{f}) \log \frac{p(I_{ir/vis}, I_{f})}{p(I_{ir/vis}) p(I_{f})},
	\end{split}
\end{equation}
where $p(I_{ir/vis}, I_{f})$ refers to the joint histogram of source images $I_{ir/vis}$ and the fused image $ I_{f}$. The $p(I_{ir/vis}$ and the $I_{f}$ compute the marginal histograms of source images and the fused image. Typically, the larger the MI value, the greater the amount of information transferred from the source images to the fusion image, and the better the fusion performance.

\textbf{VIF.} Visual Information Fidelity (VIF) measures the information fidelity of the fused image in contrast to source images. Its purpose is to calculate the distortion between the fused image and source images, which is consistent with the perception of the human visual system.


\textbf{CC.} Correlation Coefficient (CC) measures the linear correlation degree between the fused image and source images. It is mathematically defined as
\begin{equation}
	\begin{split}
		\mathrm{CC}=\frac{r(I_{ir}, I_{f}) + r(I_{vis}, I_{f})}{2},
	\end{split}	
\end{equation}
in which, $r(I_{ir}, I_{f})$ and $r(I_{vis}, I_{f})$ can be calculated by
\begin{equation}
	r(I_{s}, I_{f})=\frac{\mathbf{E}\left[\left(I_{s}-\mu(I_{s})\right) \odot\left(I_{f}-\mu(I_{f})\right)\right]}{\sqrt{\mathbf{E}\left[\left(I_{s}-\mu(I_{s})\right)^{2}\right]} \sqrt{\mathbf{E}\left[\left(I_{f}-\mu(I_{f})\right)^{2}\right]}}.	
\end{equation}
Here, $\mathbf{E}\left[\cdot \right]$ denotes the calculation of the expected value of images. $\mu(I_{s})$ and $\mu(I_{f})$ are the mean values of source image $s$ and the fused image $f$, and $\odot$ is the Hadamard product. A higher CC means that the two images are highly similar. 

Although image fusion is evaluated by more than a dozen metrics, it is reasonable to adopt the above three metrics to evaluate the fusion image generated by our IRFS, considering the total information transmission, similarity, and information fidelity that conforms to the human vision for the fused image with source images.

\begin{table*}[t]
	\centering
	\caption{Quantitative evaluations of joint thermal infrared-visible image fusion and SOD on VT5000 dataset. $\uparrow$/$\downarrow$ for a metric denotes that a larger/smaller value is better. The best results are bolded and the second-best results are highlighted in \underline{underline}.} 
	\label{tab:vt5000_fuse_sod}
	\renewcommand{\arraystretch}{0.9}
	\renewcommand{\tabcolsep}{2.4mm}
	\resizebox{1.0\linewidth}{!}{
	\begin{tabular}{rrcccccccccc}
		\toprule
		\multirow{1}{*}{Method}\centering & \multirow{1}{*}{Metric}\centering  & \textbf{\small FGAN} & \textbf{\small DIDFuse} & \textbf{\small PMGI} & \textbf{\small MFEIF} & \textbf{\small RFN} & \textbf{\small U2F} & \textbf{\small DDcGAN} & \textbf{\small GANMcC} & \textbf{\small UMF} & \textbf{\small IRFS}\\
		\hline
		\specialrule{0em}{1pt}{1pt}
		\multirow{3}{*}{\textit{}}
            
		\multirow{2}{*}{\textbf{Fusion}}
		& \text{MI}$\uparrow$      & 1.799 & 1.692 & 1.893 & 1.769 & \underline{1.962} & 1.780 & 1.798 & 1.743 & 1.741 & \textbf{2.064}  \\
		& \text{VIF}$\uparrow$     & 0.648 & 0.634 & 0.648 & \underline{1.046} & 0.728 & 0.744 & 0.930 & 0.755 & 0.981 & \textbf{1.525}  \\
  	    & \text{CC}$\uparrow$      & 1.233 & 1.378 & 1.311 & 1.368 & 1.399 & 1.383 & \underline{1.419} & 1.377 & 1.391 & \textbf{1.450}  \\
		\midrule
		\multirow{5}{*}{\textbf{+ CTDNet}}
		& $S_{\alpha}\uparrow$      & 0.847 & 0.853 & 0.852 & \underline{0.857} & \underline{0.857} & 0.853 & 0.846 & 0.854 & 0.824 & \textbf{0.871}  \\
		& $F_{\beta}\uparrow$       & 0.785 & 0.802 & 0.797 & \underline{0.805} & 0.803 & 0.799 & 0.789 & 0.800 & 0.789 & \textbf{0.835}  \\
            & $E_{\xi}\uparrow$         & 0.887 & 0.898 & 0.893 & \underline{0.898} & \underline{0.898} & 0.893 & 0.887 & 0.897 & 0.899 & \textbf{0.927}  \\
            & $\mathcal{M}\downarrow$   & 0.051 & \underline{0.046} & 0.048 & \underline{0.046} & \underline{0.046} & 0.049 & 0.051 & 0.047 & 0.049 & \textbf{0.036}  \\
            \midrule
		\multirow{5}{*}{\textbf{+ FGC$^2$Net}}
		& $S_{\alpha}\uparrow$      & 0.874 & 0.870 & 0.865 & 0.870 & 0.866 & \underline{0.872} & 0.859 & 0.867 & 0.873 & \textbf{0.877}  \\
		& $F_{\beta}\uparrow$       & 0.816 & 0.798 & 0.808 & \underline{0.818} & 0.813 & 0.815 & 0.792 & 0.801 & 0.831 & \textbf{0.835}  \\
            & $E_{\xi}\uparrow$         & 0.913 & 0.902 & 0.909 & \underline{0.915} & 0.914 & 0.909 & 0.904 & 0.904 & 0.917 & \textbf{0.922} \\
            & $\mathcal{M}\downarrow$   & \underline{0.036} & 0.039 & 0.037 & \underline{0.036} & 0.037 & 0.038 & 0.041 & 0.038 & 0.035 & \textbf{0.034} \\
		
		\bottomrule
	\end{tabular}}
\end{table*}

\subsection{Joint Image Fusion and SOD Evaluation}

We evaluate the joint multimodal image fusion and SOD performance of our IRFS on the VT5000 dataset. The state-of-the-arts image fusion methods, i.e., FGAN~\cite{FGAN}, DIDFuse~\cite{DIDFuse_2020},  PMGI~\cite{PMGI}, MFEIF~\cite{MFEIF}, RFN~\cite{RFN}, U2F~\cite{U2Fusion}, DDcGAN~\cite{DDcGAN}, GANMcC~\cite{GANMcC}, and UMF~\cite{UMF} are employed for comparison. 
To comprehensively evaluate the effectiveness of our IRFS, we combine the aforementioned fusion models with the proposed FGC$^{2}$Net and a recent SOD method CTDNet~\cite{ctdnet_21} to form some temporary multi-task frameworks.
CTDNet is a mono-modal SOD method aimed at RGB images. Accordingly, we exclusively take the fusion result of each fusion method as the input of CTDNet to perform joint fusion and SOD learning on the VT5000 dataset. To be fair, we keep the original settings of CTDNet unchanged throughout the training process.

For quantitative comparisons, Table~\ref{tab:vt5000_fuse_sod} reports the intermediate fusion results and final SOD results on the VT5000 dataset, which are obtained through temporary multi-task frameworks consisting of the aforementioned fusion models with CTDNet and our FGC$^{2}$Net. 
By comparison, our IRFS consistently performs more favorably than existing SOTA methods both on common image fusion metrics (i.e., MI, VIF, and CC) and SOD metrics (i.e., $S_{\alpha}$, $F_{\beta}$, $F_{\beta}^{w}$, and MAE).
Concretely, our IRFS outperforms the second-best methods by 45.8\% in the VIF metric. In MI and CC metrics, our IRFS also gains 5.2\% and 2.18\% improvements compared with the second place, respectively.
In contrast to the temporary multi-task frameworks formed by the fusion models and CTDNet, the proposed IRFS performs well, and the final predicted saliency maps rank first for four commonly-used metrics in the field of SOD. 
In contrast to the other temporary multi-task frameworks formed with the FGC$^{2}$Net, our IRFS is still able to rank first. Compared with the second-best method, the gain reaches 2.08\% in mean $F_{\beta}$, 5.56\% in mean MAE score.
In addition, we observe that these temporary multi-task frameworks formed with the FGC$^{2}$Net perform better than those formed with the recent CTDNet~\cite{ctdnet_21}.
Analysis of the above results quantitatively illustrates the superiority of the proposed IRFS paradigm.

\begin{figure*}[h]
	\centering	
	\subfigure{
        \rotatebox{90}{\scriptsize{~~\bf Fusion}}
		\begin{minipage}[t]{0.0850\linewidth}
			\centering
			\includegraphics[width=1\linewidth]{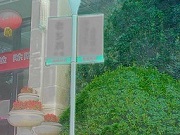}
		\end{minipage}
	} \hspace{-3.6mm}
	\subfigure{
		\begin{minipage}[t]{0.0850\linewidth}
			\centering
			\includegraphics[width=1\linewidth]{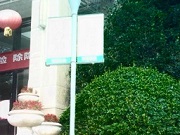}
		\end{minipage}
	} \hspace{-3.6mm}
         \subfigure{
            \begin{minipage}[t]{0.0850\linewidth}
                \centering
                \includegraphics[width=1\linewidth]{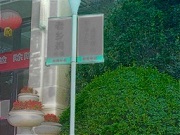}
            \end{minipage}
        } \hspace{-3.6mm}
        \subfigure{
            \begin{minipage}[t]{0.0850\linewidth}
                \centering
                \includegraphics[width=1\linewidth]{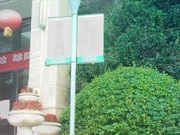}
            \end{minipage}
        } \hspace{-3.6mm}
        \subfigure{
            \begin{minipage}[t]{0.0850\linewidth}
                \centering
                \includegraphics[width=1\linewidth]{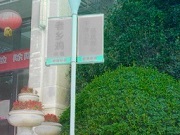}
            \end{minipage}
        } \hspace{-3.6mm}
        \subfigure{
            \begin{minipage}[t]{0.0850\linewidth}
                \centering
                \includegraphics[width=1\linewidth]{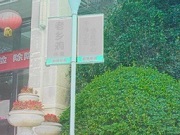}
            \end{minipage}
        } \hspace{-3.6mm}
        \subfigure{
            \begin{minipage}[t]{0.0850\linewidth}
                \centering
                \includegraphics[width=1\linewidth]{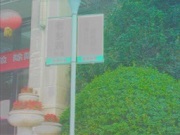}
            \end{minipage}
        } \hspace{-3.6mm}
        \subfigure{
            \begin{minipage}[t]{0.0850\linewidth}
                \centering
                \includegraphics[width=1\linewidth]{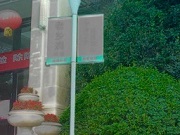}
            \end{minipage}
        } \hspace{-3.6mm}
        \subfigure{
            \begin{minipage}[t]{0.0850\linewidth}
                \centering
                \includegraphics[width=1\linewidth]{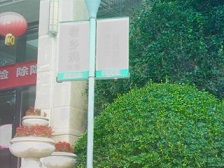}
            \end{minipage}
        } \hspace{-3.6mm}
        \subfigure{
            \begin{minipage}[t]{0.0850\linewidth}
                \centering
                \includegraphics[width=1\linewidth]{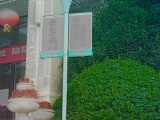}
            \end{minipage}
        } \hspace{-3.6mm}
        \subfigure{
            \begin{minipage}[t]{0.0850\linewidth}
                \centering
                \includegraphics[width=1\linewidth]{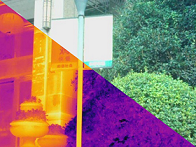}
            \end{minipage}
        }\\
        \vspace{-2mm}

        \subfigure{
		\rotatebox{90}{\scriptsize{\bf ~~~CTDNet}}
		\begin{minipage}[t]{0.0850\linewidth}
			\centering
			\includegraphics[width=1\linewidth]{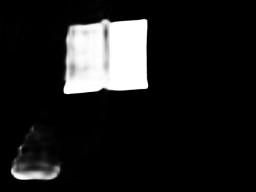}
		\end{minipage}
	} \hspace{-3.6mm}
	\subfigure{
		\begin{minipage}[t]{0.0850\linewidth}
			\centering
			\includegraphics[width=1\linewidth]{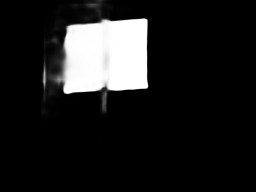}
		\end{minipage}
	} \hspace{-3.6mm}
        \subfigure{
		\begin{minipage}[t]{0.0850\linewidth}
			\centering
			\includegraphics[width=1\linewidth]{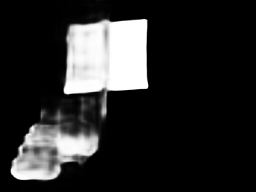}
		\end{minipage}
	} \hspace{-3.6mm}
        \subfigure{
		\begin{minipage}[t]{0.0850\linewidth}
			\centering
			\includegraphics[width=1\linewidth]{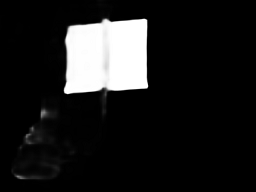}
		\end{minipage}
	} \hspace{-3.6mm}
        \subfigure{
		\begin{minipage}[t]{0.0850\linewidth}
			\centering
			\includegraphics[width=1\linewidth]{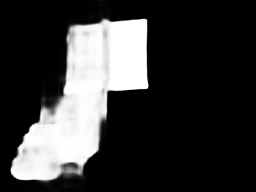}
		\end{minipage}
	} \hspace{-3.6mm}
        \subfigure{
		\begin{minipage}[t]{0.0850\linewidth}
			\centering
			\includegraphics[width=1\linewidth]{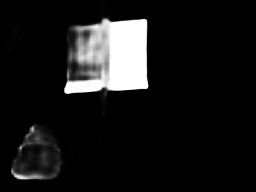}
		\end{minipage}
	} \hspace{-3.6mm}
        \subfigure{
		\begin{minipage}[t]{0.0850\linewidth}
			\centering
			\includegraphics[width=1\linewidth]{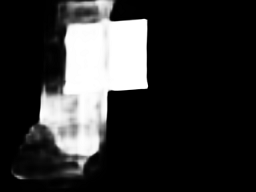}
		\end{minipage}
	} \hspace{-3.6mm}
        \subfigure{
		\begin{minipage}[t]{0.0850\linewidth}
			\centering
			\includegraphics[width=1\linewidth]{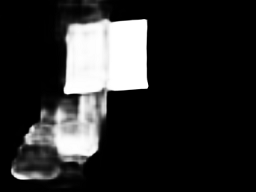}
		\end{minipage}
	} \hspace{-3.6mm}
        \subfigure{
		\begin{minipage}[t]{0.0850\linewidth}
			\centering
			\includegraphics[width=1\linewidth]{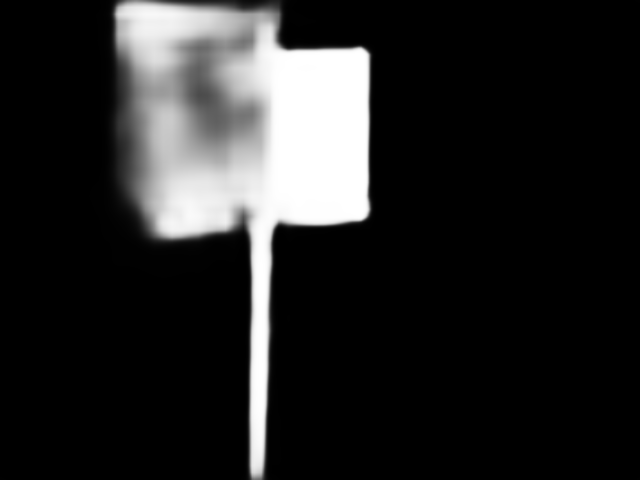}
		\end{minipage}
	} \hspace{-3.6mm}
        \subfigure{
		\begin{minipage}[t]{0.0850\linewidth}
			\centering
			\includegraphics[width=1\linewidth]{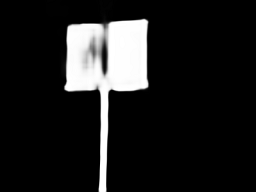}
		\end{minipage}
	} \hspace{-3.6mm}
        \subfigure{
		\begin{minipage}[t]{0.0850\linewidth}
			\centering
			\includegraphics[width=1\linewidth]{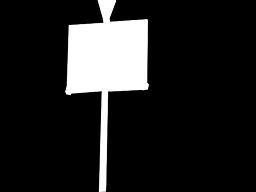}
		\end{minipage}
	}\\
        \vspace{-2mm}
        
        \subfigure{
		\rotatebox{90}{\scriptsize{\bf ~~FGC$^2$Net}}
		\begin{minipage}[t]{0.0850\linewidth}
			\centering
			\includegraphics[width=1\linewidth]{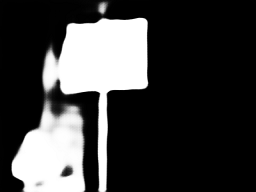}\\
                \footnotesize FGAN
		\end{minipage}
	} \hspace{-3.6mm}
	\subfigure{
		\begin{minipage}[t]{0.0850\linewidth}
			\centering
			\includegraphics[width=1\linewidth]{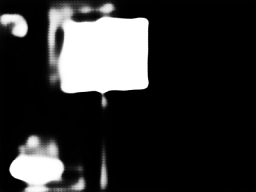}\\
                \footnotesize DIDFuse
		\end{minipage}
	} \hspace{-3.6mm}
        \subfigure{
		\begin{minipage}[t]{0.0850\linewidth}
			\centering
			\includegraphics[width=1\linewidth]{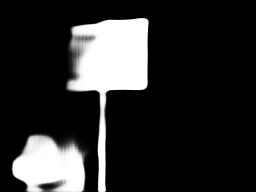}\\
                \footnotesize PMGI
		\end{minipage}
	} \hspace{-3.6mm}
        \subfigure{
		\begin{minipage}[t]{0.0850\linewidth}
			\centering
			\includegraphics[width=1\linewidth]{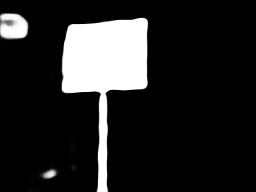}\\
                \footnotesize MFEIF
		\end{minipage}
	} \hspace{-3.6mm}
        \subfigure{
		\begin{minipage}[t]{0.0850\linewidth}
			\centering
			\includegraphics[width=1\linewidth]{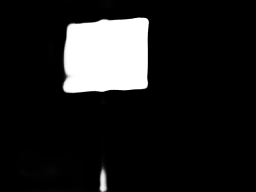}\\
                \footnotesize RFN
		\end{minipage}
	} \hspace{-3.6mm}
        \subfigure{
		\begin{minipage}[t]{0.0850\linewidth}
			\centering
			\includegraphics[width=1\linewidth]{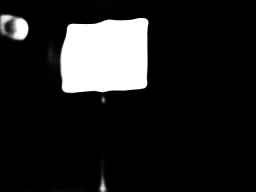}\\
                \footnotesize U2F
		\end{minipage}
	} \hspace{-3.6mm}
        \subfigure{
		\begin{minipage}[t]{0.0850\linewidth}
			\centering
			\includegraphics[width=1\linewidth]{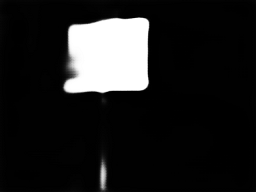}\\
                \footnotesize DDcGAN
		\end{minipage}
	} \hspace{-3.6mm}
        \subfigure{
		\begin{minipage}[t]{0.0850\linewidth}
			\centering
			\includegraphics[width=1\linewidth]{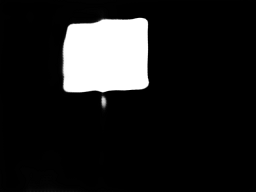}\\
                \footnotesize GANMcC
		\end{minipage}
	} \hspace{-3.6mm}
        \subfigure{
		\begin{minipage}[t]{0.0850\linewidth}
			\centering
			\includegraphics[width=1\linewidth]{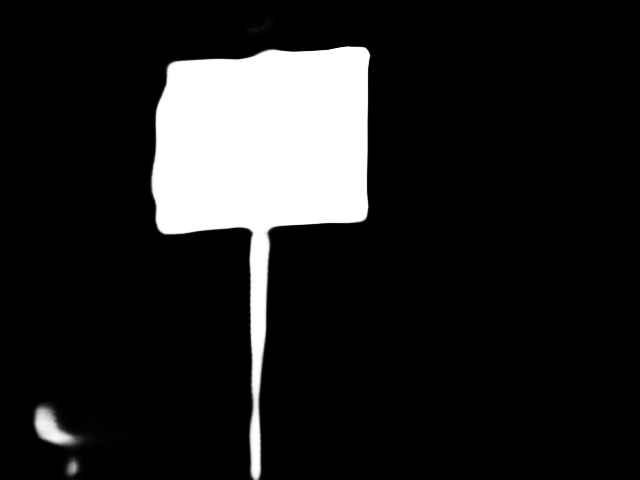}\\
                \footnotesize UMF
		\end{minipage}
	} \hspace{-3.6mm}
        \subfigure{
		\begin{minipage}[t]{0.0850\linewidth}
			\centering
			\includegraphics[width=1\linewidth]{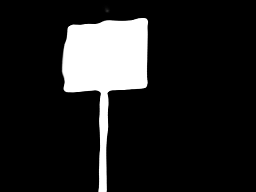}\\
                \footnotesize IRFS
		\end{minipage}
	} \hspace{-3.6mm}
        \subfigure{
		\begin{minipage}[t]{0.0850\linewidth}
			\centering
			\includegraphics[width=1\linewidth]{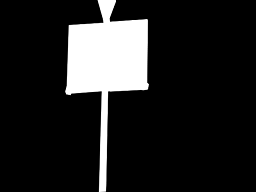}\\
                \footnotesize GT
		\end{minipage}
	} \vspace{-2mm}

	\subfigure{
        \rotatebox{90}{\scriptsize{~~\bf Fusion}}
		\begin{minipage}[t]{0.0850\linewidth}
			\centering
			\includegraphics[width=1\linewidth]{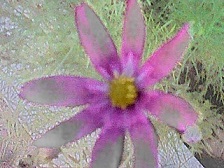}
		\end{minipage}
	} \hspace{-3.6mm}
	\subfigure{
		\begin{minipage}[t]{0.0850\linewidth}
			\centering
			\includegraphics[width=1\linewidth]{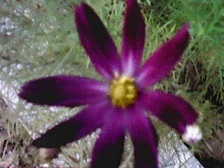}
		\end{minipage}
	} \hspace{-3.6mm}
         \subfigure{
            \begin{minipage}[t]{0.0850\linewidth}
                \centering
                \includegraphics[width=1\linewidth]{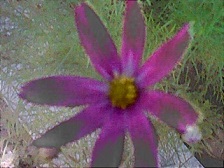}
            \end{minipage}
        } \hspace{-3.6mm}
        \subfigure{
            \begin{minipage}[t]{0.0850\linewidth}
                \centering
                \includegraphics[width=1\linewidth]{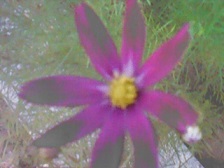}
            \end{minipage}
        } \hspace{-3.6mm}
        \subfigure{
            \begin{minipage}[t]{0.0850\linewidth}
                \centering
                \includegraphics[width=1\linewidth]{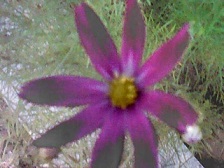}
            \end{minipage}
        } \hspace{-3.6mm}
        \subfigure{
            \begin{minipage}[t]{0.0850\linewidth}
                \centering
                \includegraphics[width=1\linewidth]{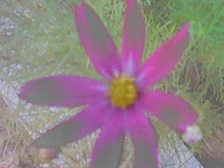}
            \end{minipage}
        } \hspace{-3.6mm}
        \subfigure{
            \begin{minipage}[t]{0.0850\linewidth}
                \centering
                \includegraphics[width=1\linewidth]{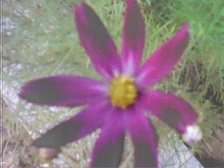}
            \end{minipage}
        } \hspace{-3.6mm}
        \subfigure{
            \begin{minipage}[t]{0.0850\linewidth}
                \centering
                \includegraphics[width=1\linewidth]{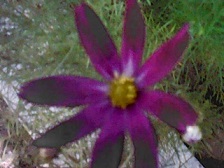}
            \end{minipage}
        } \hspace{-3.6mm}
        \subfigure{
            \begin{minipage}[t]{0.0850\linewidth}
                \centering
                \includegraphics[width=1\linewidth]{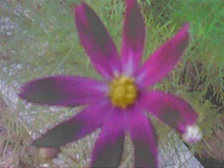}
            \end{minipage}
        } \hspace{-3.6mm}
        \subfigure{
            \begin{minipage}[t]{0.0850\linewidth}
                \centering
                \includegraphics[width=1\linewidth]{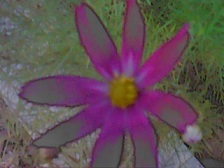}
            \end{minipage}
        } \hspace{-3.6mm}
        \subfigure{
            \begin{minipage}[t]{0.0850\linewidth}
                \centering
                \includegraphics[width=1\linewidth]{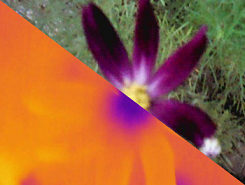}
            \end{minipage}
        }\\
        \vspace{-2mm}

        \subfigure{
		\rotatebox{90}{\scriptsize{\bf ~~~CTDNet}}
		\begin{minipage}[t]{0.0850\linewidth}
			\centering
			\includegraphics[width=1\linewidth]{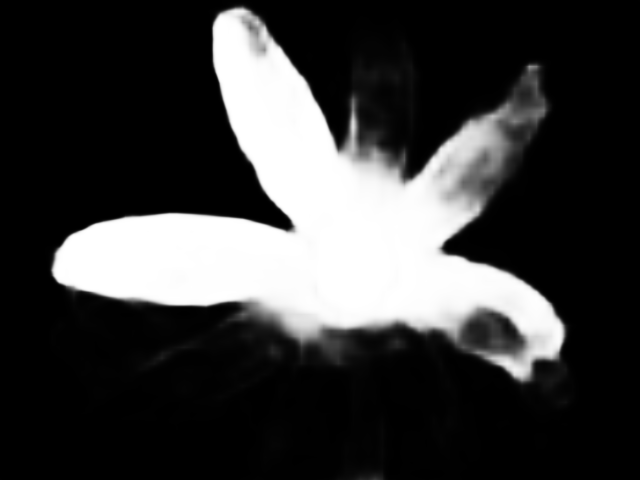}
		\end{minipage}
	} \hspace{-3.6mm}
	\subfigure{
		\begin{minipage}[t]{0.0850\linewidth}
			\centering
			\includegraphics[width=1\linewidth]{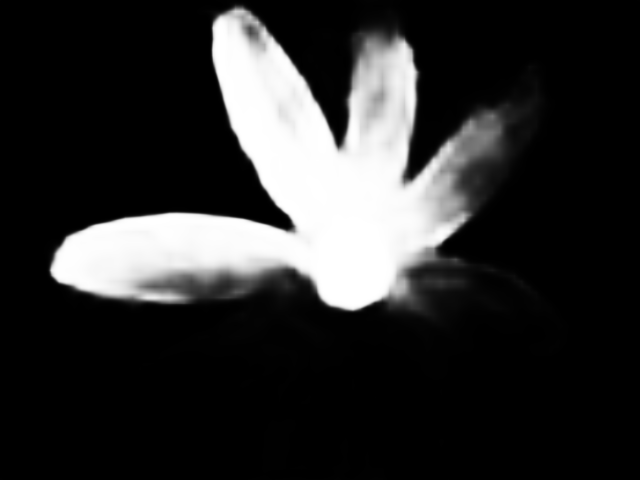}
		\end{minipage}
	} \hspace{-3.6mm}
        \subfigure{
		\begin{minipage}[t]{0.0850\linewidth}
			\centering
			\includegraphics[width=1\linewidth]{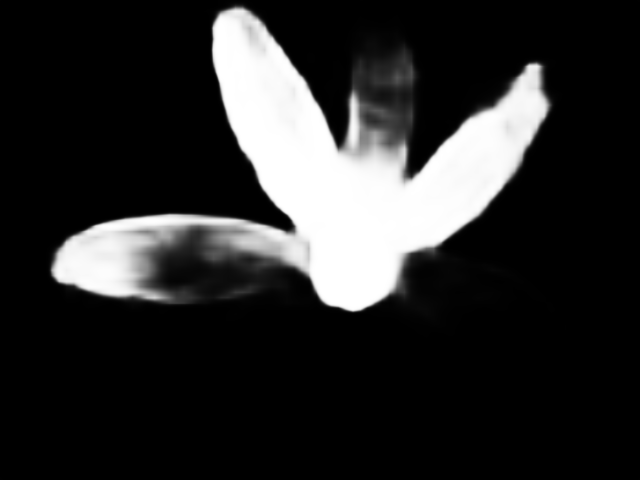}
		\end{minipage}
	} \hspace{-3.6mm}
        \subfigure{
		\begin{minipage}[t]{0.0850\linewidth}
			\centering
			\includegraphics[width=1\linewidth]{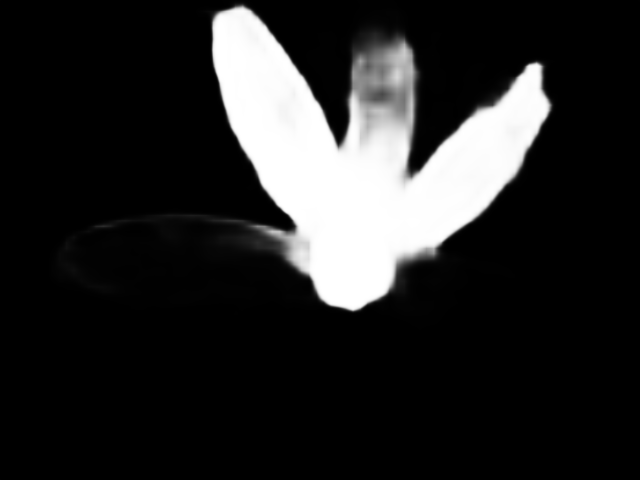}
		\end{minipage}
	} \hspace{-3.6mm}
        \subfigure{
		\begin{minipage}[t]{0.0850\linewidth}
			\centering
			\includegraphics[width=1\linewidth]{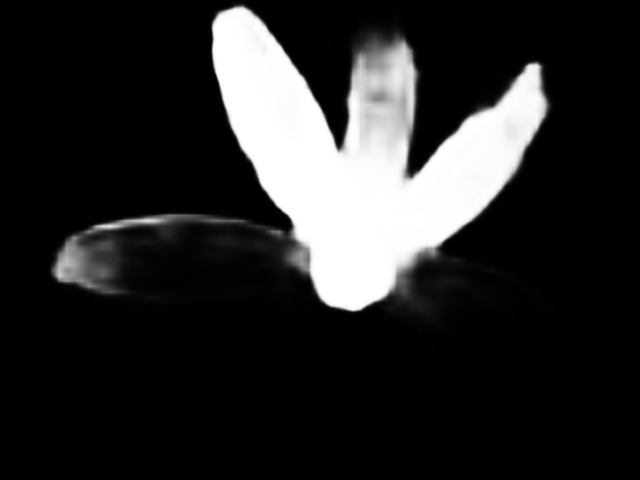}
		\end{minipage}
	} \hspace{-3.6mm}
        \subfigure{
		\begin{minipage}[t]{0.0850\linewidth}
			\centering
			\includegraphics[width=1\linewidth]{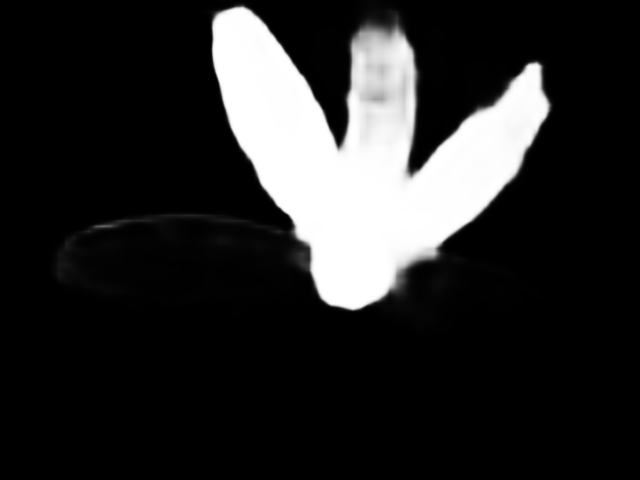}
		\end{minipage}
	} \hspace{-3.6mm}
        \subfigure{
		\begin{minipage}[t]{0.0850\linewidth}
			\centering
			\includegraphics[width=1\linewidth]{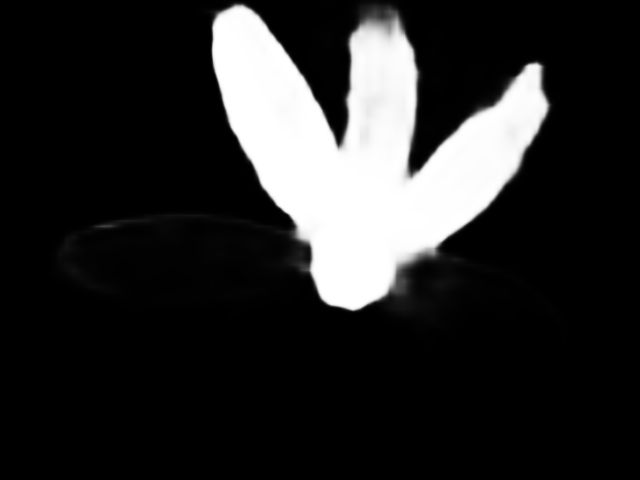}
		\end{minipage}
	} \hspace{-3.6mm}
        \subfigure{
		\begin{minipage}[t]{0.0850\linewidth}
			\centering
			\includegraphics[width=1\linewidth]{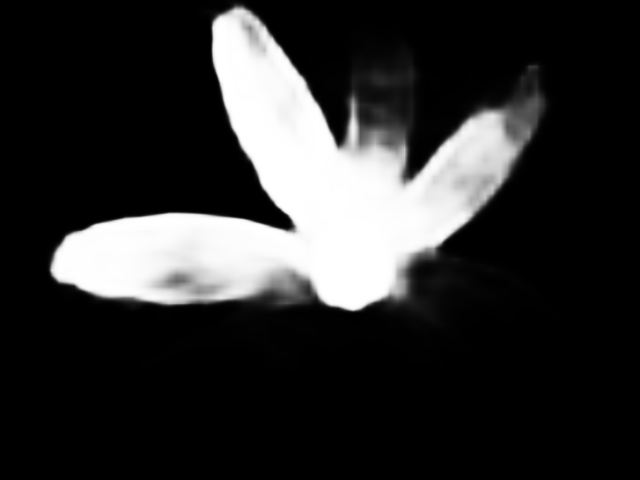}
		\end{minipage}
	} \hspace{-3.6mm}
        \subfigure{
		\begin{minipage}[t]{0.0850\linewidth}
			\centering
			\includegraphics[width=1\linewidth]{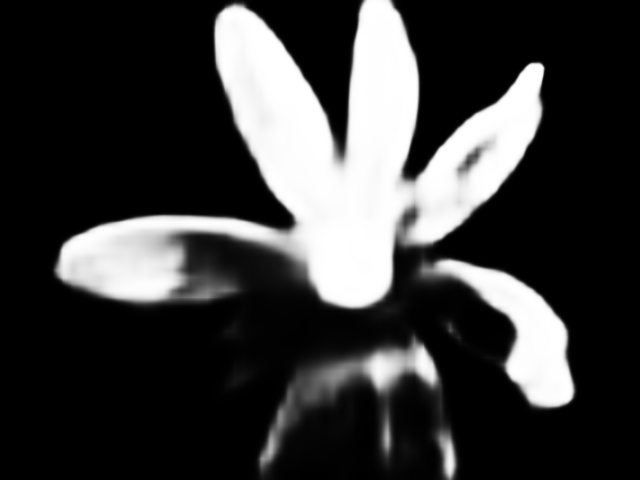}
		\end{minipage}
	} \hspace{-3.6mm}
        \subfigure{
		\begin{minipage}[t]{0.0850\linewidth}
			\centering
			\includegraphics[width=1\linewidth]{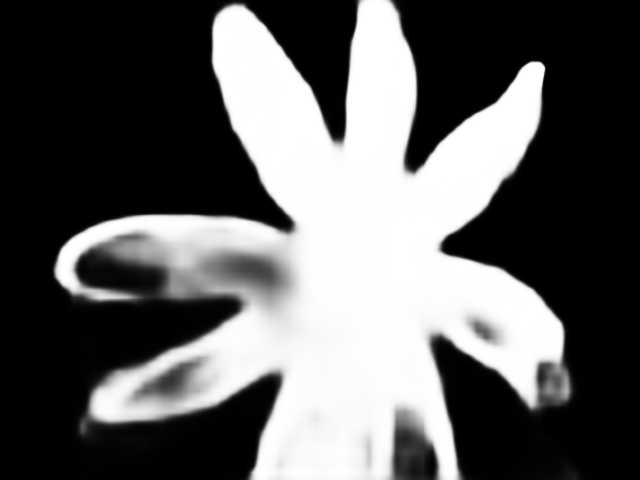}
		\end{minipage}
	} \hspace{-3.6mm}
        \subfigure{
		\begin{minipage}[t]{0.0850\linewidth}
			\centering
			\includegraphics[width=1\linewidth]{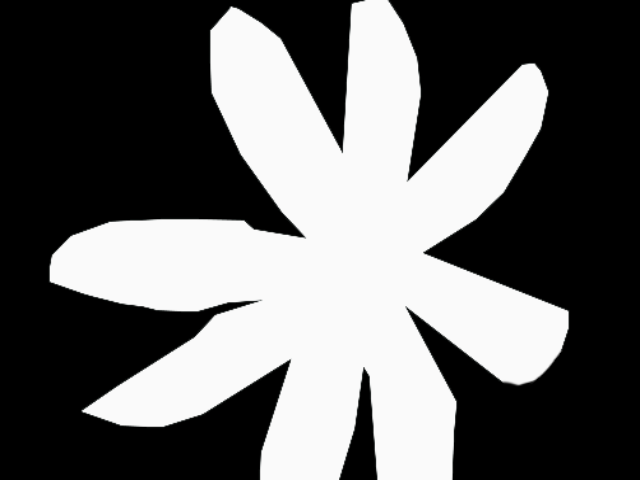}
		\end{minipage}
	}\\
        \vspace{-2mm}
        
        \subfigure{
		\rotatebox{90}{\scriptsize{\bf ~~FGC$^2$Net}}
		\begin{minipage}[t]{0.0850\linewidth}
			\centering
			\includegraphics[width=1\linewidth]{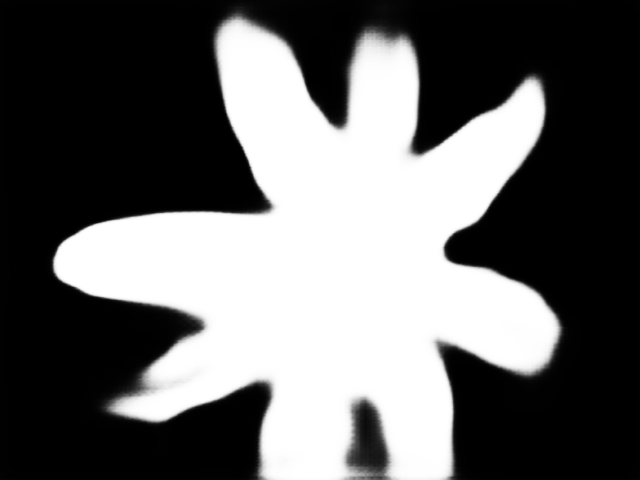}\\
                \footnotesize FGAN
		\end{minipage}
	} \hspace{-3.6mm}
	\subfigure{
		\begin{minipage}[t]{0.0850\linewidth}
			\centering
			\includegraphics[width=1\linewidth]{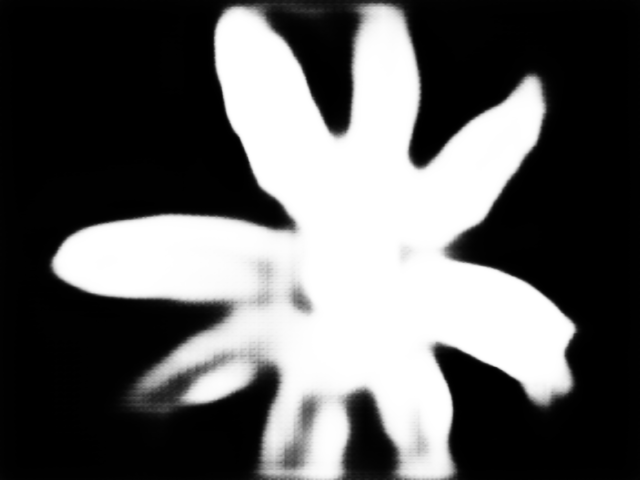}\\
                \footnotesize DIDFuse
		\end{minipage}
	} \hspace{-3.6mm}
        \subfigure{
		\begin{minipage}[t]{0.0850\linewidth}
			\centering
			\includegraphics[width=1\linewidth]{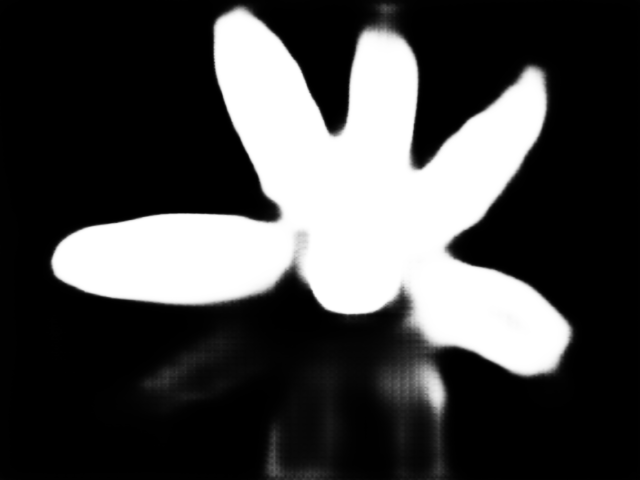}\\
                \footnotesize PMGI
		\end{minipage}
	} \hspace{-3.6mm}
        \subfigure{
		\begin{minipage}[t]{0.0850\linewidth}
			\centering
			\includegraphics[width=1\linewidth]{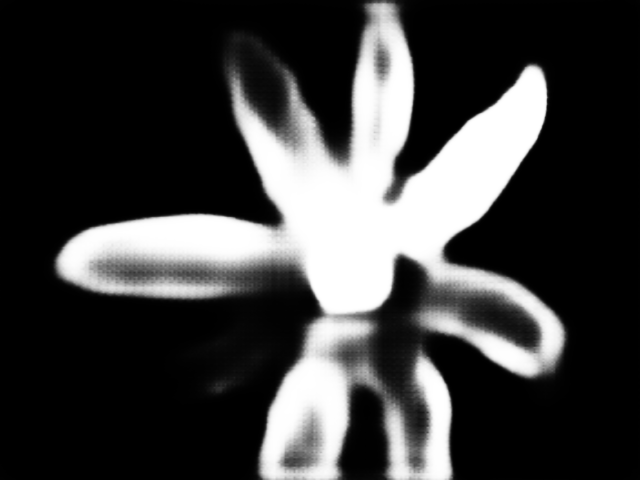}\\
                \footnotesize MFEIF
		\end{minipage}
	} \hspace{-3.6mm}
        \subfigure{
		\begin{minipage}[t]{0.0850\linewidth}
			\centering
			\includegraphics[width=1\linewidth]{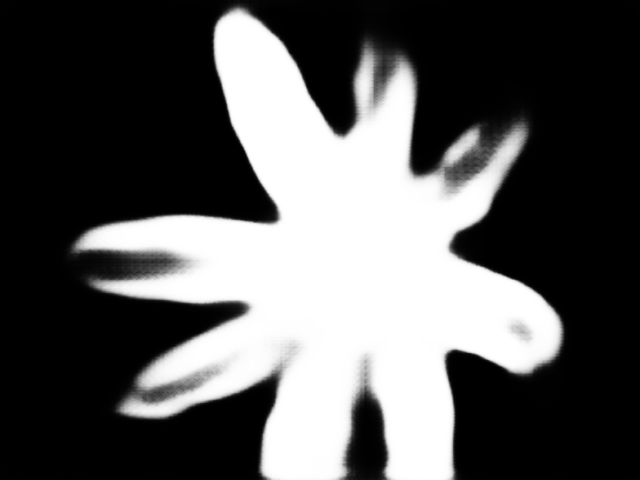}\\
                \footnotesize RFN
		\end{minipage}
	} \hspace{-3.6mm}
        \subfigure{
		\begin{minipage}[t]{0.0850\linewidth}
			\centering
			\includegraphics[width=1\linewidth]{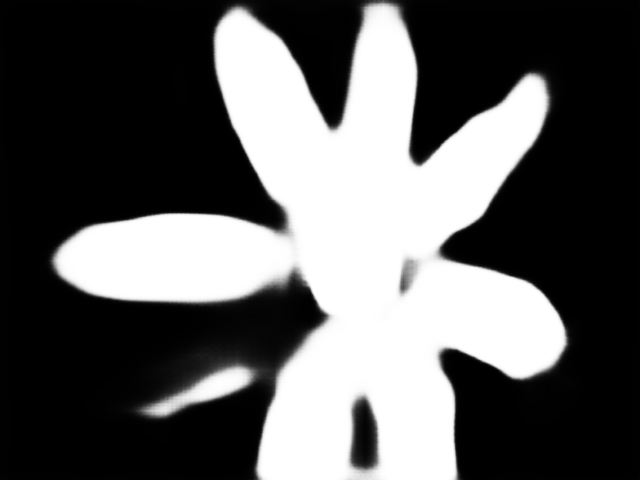}\\
                \footnotesize U2F
		\end{minipage}
	} \hspace{-3.6mm}
        \subfigure{
		\begin{minipage}[t]{0.0850\linewidth}
			\centering
			\includegraphics[width=1\linewidth]{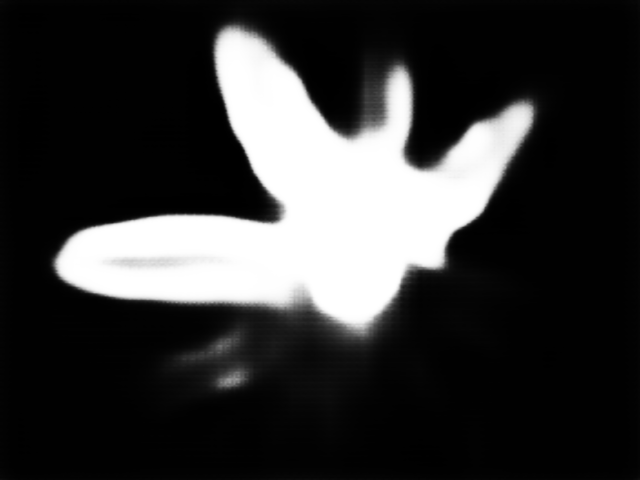}\\
                \footnotesize DDcGAN
		\end{minipage}
	} \hspace{-3.6mm}
        \subfigure{
		\begin{minipage}[t]{0.0850\linewidth}
			\centering
			\includegraphics[width=1\linewidth]{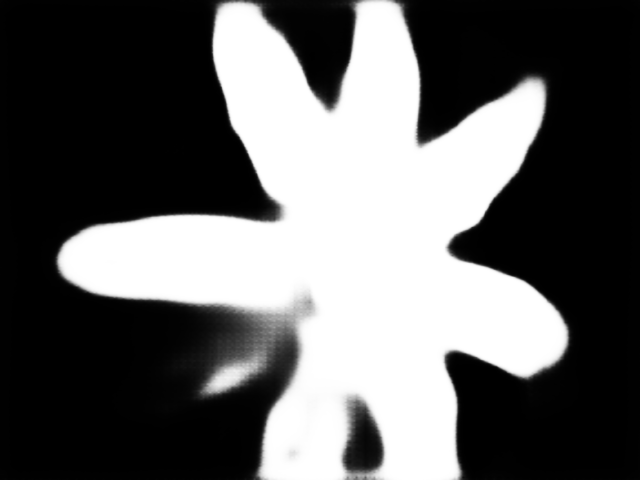}\\
                \footnotesize GANMcC
		\end{minipage}
	} \hspace{-3.6mm}
        \subfigure{
		\begin{minipage}[t]{0.0850\linewidth}
			\centering
			\includegraphics[width=1\linewidth]{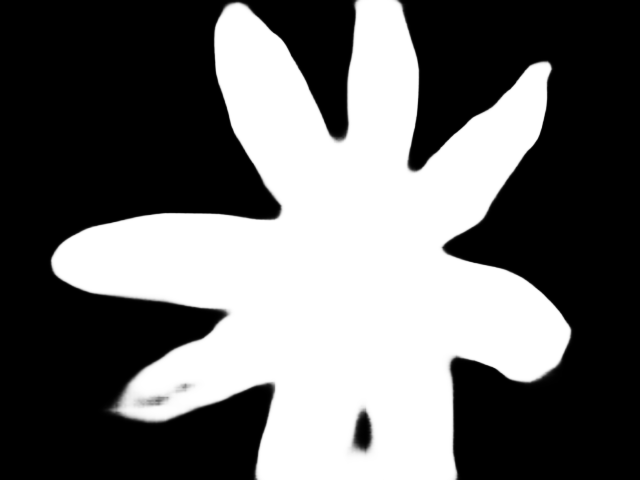}\\
                \footnotesize UMF
		\end{minipage}
	} \hspace{-3.6mm}
        \subfigure{
		\begin{minipage}[t]{0.0850\linewidth}
			\centering
			\includegraphics[width=1\linewidth]{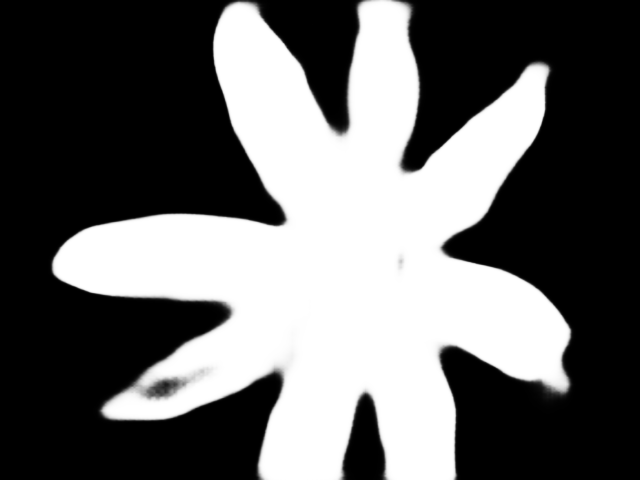}\\
                \footnotesize IRFS
		\end{minipage}
	} \hspace{-3.6mm}
        \subfigure{
		\begin{minipage}[t]{0.0850\linewidth}
			\centering
			\includegraphics[width=1\linewidth]{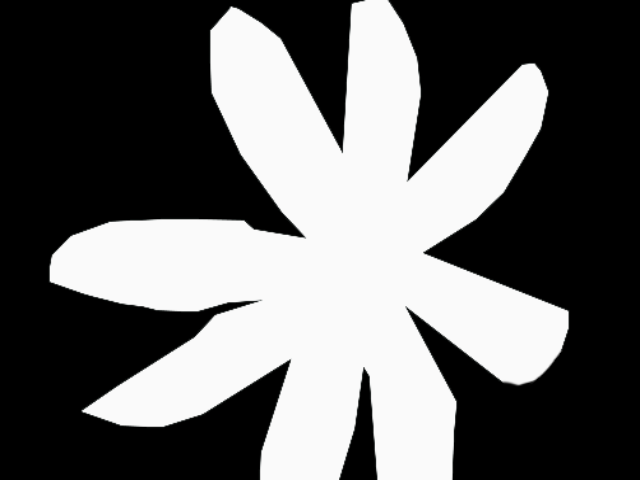}\\
                \footnotesize GT
		\end{minipage}
	}
	\caption{Qualitative evaluations of joint infrared-visible image fusion and SOD on VT5000 dataset. The first row refers to the fusion results, while the second and last rows refer to the saliency maps predicted by our FGC$^2$Net and CTDNet~\cite{ctdnet_21}, respectively.}
\label{fig:joint_fusion_sod}
        \vspace{-4mm}
\end{figure*}

For qualitative results, we show two examples in Figure~\ref{fig:joint_fusion_sod}. In each example, a group of fused images generated from existing fusion models, as well as two groups of corresponding saliency maps derived from CTDNet~\cite{ctdnet_21} and the FGC$^{2}$Net of our IRFS framework are exhibited.
By comparing these results, we can find that the contrast difference between the object and the background is more pronounced in fused images generated by our IRFS, and overexposure is suppressed, which contributes to more accurate predictions of the saliency maps than other temporary multi-task frameworks, as shown in Figure~\ref{fig:joint_fusion_sod}.

Quantitative and qualitative results indicated that, on the one hand, a tightly cooperative relationship between thermal infrared-visible image fusion and SOD is existing. On the other hand, the effectiveness of our interactively reinforced paradigm for joint image fusion and SOD is supported.

\begin{table*}[t]
	\centering
	\caption{Quantitative fusion results. $\uparrow$/$\downarrow$ for a metric denotes that a larger/smaller value is better. The best results are bolded and the second-best results are highlighted in \underline{underline}.} 
	\label{tab:fusion}
	
	\renewcommand{\arraystretch}{0.9}
	\renewcommand{\tabcolsep}{2.4mm}
	\resizebox{1.0\linewidth}{!}{
	\begin{tabular}{lr|ccccccccc|c}
		\toprule
		&\multirow{1}{*}{Metric}\centering  & \textbf{\small FGAN}   & \textbf{\small DIDFuse} & \textbf{\small PMGI} & \textbf{\small MFEIF} & \textbf{\small RFN} & \textbf{\small U2Fusion} & \textbf{\small DDcGAN} & \textbf{\small GANMcC}  & \textbf{\small UMF} &\textbf{\small IRFS}\\
		\hline
		\specialrule{0em}{1pt}{1pt}
		& Params.~(Mb)$\downarrow$ &$0.074$	&$0.261$		&$\textbf{0.042}$	&$0.158$	&$10.936$ &$0.659$	&$1.098$    &$1.864$  & $0.80$ &$\underline{0.06}$\\
		\hline
		\specialrule{0em}{1pt}{1pt}
		\multirow{3}{*}{\textit{}}
		\multirow{3}{*}{\textbf{TNO}}
		& $\text{MI}\uparrow$      & 1.621 & 1.888 & \underline{1.920} & \textbf{1.968} & 1.671 & 1.618 & 1.468 & 1.796 & 1.818 & 1.843 \\
		& $\text{VIF}\uparrow$     & 0.798 & 0.778 & 0.779 & \underline{1.126} & 0.836 & 0.825 & 0.567 & 0.914 & 1.078 & \textbf{1.230} \\
		& $\text{CC}\uparrow$      & 1.213 & 1.306 & 1.372 & 1.349 & 1.427 & \underline{1.430} & 1.258 & 1.413 & 1.383 & \textbf{1.456} \\
		\midrule
		\multirow{3}{*}{\textbf{Road}}
		& $\text{MI}\uparrow$      & 2.047 & 1.925 & \underline{2.269} & 2.170 & 1.999 & 1.812 & 1.723 & 1.876 & 2.019 & \textbf{2.472} \\
		& $\text{VIF}\uparrow$     & 0.703 & 0.881 & 0.874 & \underline{1.164} & 1.043 & 0.768 & 0.610 & 0.952 & 1.141 & \textbf{1.173} \\
		& $\text{CC}\uparrow$      & 1.438 & 1.533 & 1.482 & 1.595 & \underline{1.613} & 1.602 & 1.414 & 1.573 & 1.605 & \textbf{1.616}\\
            \midrule
		\multirow{3}{*}{\textbf{M$^3$FD}}
		& $\text{MI}\uparrow$  & 1.589 & 1.803 & 1.948 & 1.574 & 1.877 & 1.676 & 1.333 & 1.620 & \underline{1.960} & \textbf{1.999} \\
		& $\text{VIF}\uparrow$ & 0.800 & 0.695 & 0.783 & \underline{1.140} & 0.907 & 0.819 & 0.444 & 0.944 & 1.100 & \textbf{1.158} \\
		& $\text{CC}\uparrow$  & 0.774 & 0.794 & 0.920 & 0.834 & 0.851 & 0.893 & 0.831 & 0.936 & \underline{0.939} & \textbf{0.967} \\
		\bottomrule
	\end{tabular}}
	\vspace{-2mm}
\end{table*}

\begin{figure*}[t]
	\begin{center}
		\resizebox{1.0\linewidth}{!}{
		\begin{tabular}{cccc}	
			
			\includegraphics[width = 0.195\linewidth ,height=0.066\textheight]{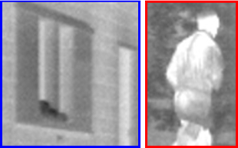} & 
			\hspace{-0.46cm}
			\includegraphics[width = 0.195\linewidth ,height=0.066\textheight]{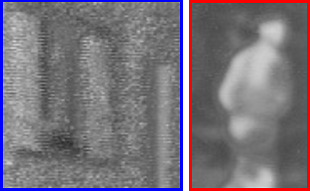}  & 
			\hspace{-0.46cm}
			\includegraphics[width = 0.195\linewidth ,height=0.066\textheight]{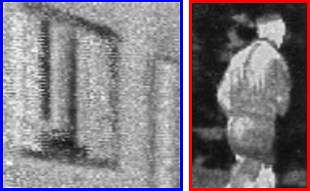}& 
			\hspace{-0.46cm}
			\includegraphics[width = 0.195\linewidth ,height=0.066\textheight]{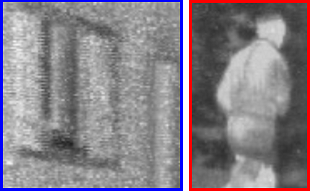}\\
                \footnotesize IR
			& \hspace{-0.46cm} \footnotesize FGAN~\cite{FGAN}
			& \hspace{-0.46cm} \footnotesize DIDFuse~\cite{DIDFuse_2020}
			& \hspace{-0.46cm} \footnotesize PMGI~\cite{PMGI} \\
   
                \includegraphics[width = 0.195\linewidth ,height=0.066\textheight]{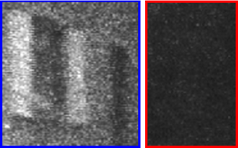} & 
			\hspace{-0.46cm}
			\includegraphics[width = 0.195\linewidth ,height=0.066\textheight]{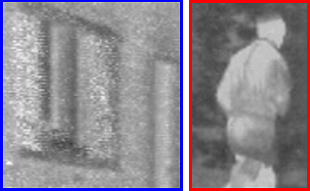} &
                \hspace{-0.46cm}
                \includegraphics[width = 0.195\linewidth ,height=0.066\textheight]{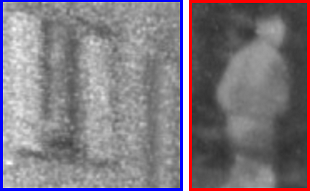}    & 
                \hspace{-0.46cm}
			\includegraphics[width = 0.195\linewidth ,height=0.066\textheight]{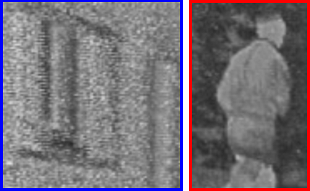} \\	
			
			\footnotesize VIS
			& \hspace{-0.46cm} \footnotesize MFEIF~\cite{MFEIF}
			& \hspace{-0.46cm} \footnotesize RFN~\cite{RFN}
			& \hspace{-0.46cm} \footnotesize U2Fusion~\cite{U2Fusion} \\
			
			\includegraphics[width = 0.195\linewidth ,height=0.066\textheight]{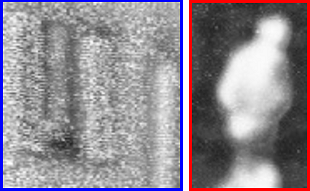} & \hspace{-0.46cm}
			\includegraphics[width = 0.195\linewidth ,height=0.066\textheight]{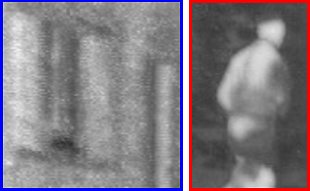} & \hspace{-0.46cm}
                \includegraphics[width = 0.195\linewidth ,height=0.066\textheight]{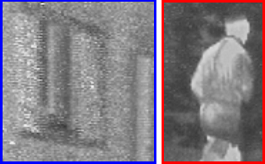} & \hspace{-0.46cm}
			\includegraphics[width = 0.195\linewidth ,height=0.066\textheight]{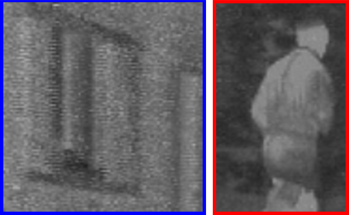}  \\						
			\footnotesize DDcGAN~\cite{DDcGAN}
			& \hspace{-0.46cm} \footnotesize GANMcC~\cite{GANMcC}
                & \hspace{-0.46cm} \footnotesize UMF~\cite{UMF}
			& \hspace{-0.46cm} \footnotesize IRFS (Ours)			
			\\
		\end{tabular}}
	\end{center}
         \vspace{-4mm}
	\caption{Qualitative fusion results of our IRFS versus state-of-the-art infrared-visible image fusion methods on TNO dataset.}
	\label{fig:fusion-tno}
	\vspace{-4mm}
\end{figure*}

\subsection{Generalization Analysis on Image Fusion}
We evaluate the intermediate fusion results of our IRFS through the comparison with existing fusion methods including FGAN~\cite{FGAN}, DIDFuse~\cite{DIDFuse_2020},  PMGI~\cite{PMGI}, MFEIF~\cite{MFEIF}, RFN~\cite{RFN}, U2F~\cite{U2Fusion}, DDcGAN~\cite{DDcGAN}, GANMcC~\cite{GANMcC}, and UMF~\cite{UMF}.

\begin{figure*}[t]
	\begin{center}
		\resizebox{1.0\linewidth}{!}{
		\begin{tabular}{cccc}	
			\includegraphics[width = 0.195\linewidth ,height=0.065\textheight]{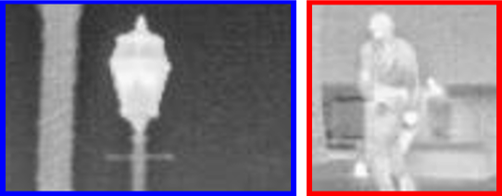} & 
			\hspace{-0.46cm}
			\includegraphics[width = 0.195\linewidth ,height=0.066\textheight]{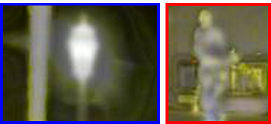}  & 
			\hspace{-0.46cm}
			\includegraphics[width = 0.195\linewidth ,height=0.066\textheight]{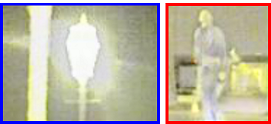}& 
			\hspace{-0.46cm}
			\includegraphics[width = 0.195\linewidth ,height=0.066\textheight]{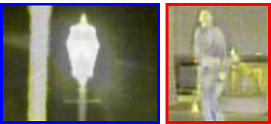} \\
                \footnotesize IR
			& \hspace{-0.46cm} \footnotesize FGAN~\cite{FGAN}
			& \hspace{-0.46cm} \footnotesize DIDFuse~\cite{DIDFuse_2020}
			& \hspace{-0.46cm} \footnotesize PMGI~\cite{PMGI} \\
   
			\includegraphics[width = 0.195\linewidth ,height=0.064\textheight]{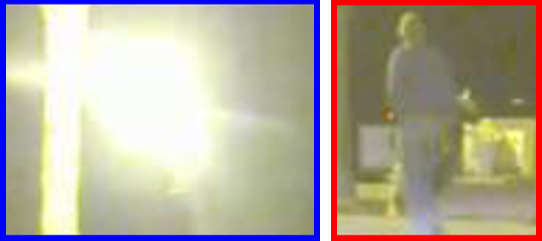} & 
                \hspace{-0.46cm}
			\includegraphics[width = 0.195\linewidth ,height=0.066\textheight]{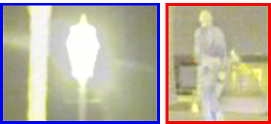} & 	
                \hspace{-0.46cm}
                \includegraphics[width = 0.195\linewidth ,height=0.066\textheight]{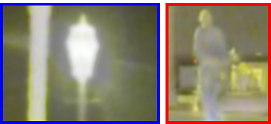}   & 
                \hspace{-0.46cm}
			\includegraphics[width = 0.195\linewidth ,height=0.066\textheight]{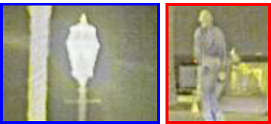} \\
			\footnotesize VIS
			& \hspace{-0.46cm} \footnotesize FGAN~\cite{MFEIF}
			& \hspace{-0.46cm} \footnotesize RFN~\cite{RFN}
			& \hspace{-0.46cm} \footnotesize U2Fusion~\cite{U2Fusion}\\
			
			\includegraphics[width = 0.195\linewidth ,height=0.066\textheight]{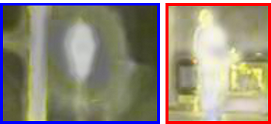} & \hspace{-0.46cm}
			\includegraphics[width = 0.195\linewidth ,height=0.066\textheight]{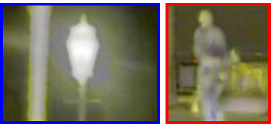} & \hspace{-0.46cm}
                \includegraphics[width = 0.195\linewidth ,height=0.064\textheight]{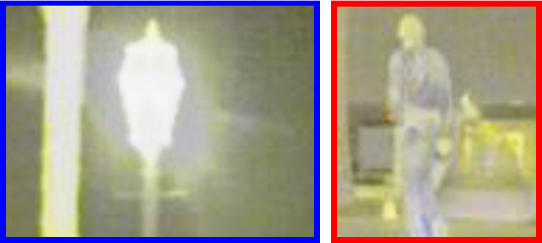} & \hspace{-0.46cm}
			\includegraphics[width = 0.195\linewidth ,height=0.066\textheight]{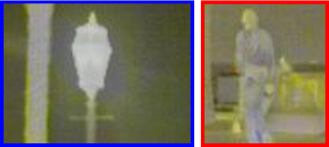}  \\						
			\footnotesize DDcGAN~\cite{DDcGAN}
			& \hspace{-0.46cm} \footnotesize GANMcC~\cite{GANMcC}
                & \hspace{-0.46cm} \footnotesize UMF~\cite{UMF}
			& \hspace{-0.46cm} \footnotesize IRFS (Ours)		
			\\
		\end{tabular}}
	\end{center}
        \vspace{-4mm}
	\caption{Qualitative fusion results of our IRFS versus state-of-the-art infrared-visible image fusion methods on RoadScene dataset.}
	\label{fig:fusion-road}
	\vspace{-4mm}
\end{figure*}

\begin{figure*}[t]
	\begin{center}
		\resizebox{1.0\linewidth}{!}{
		\begin{tabular}{cccc}	
			
			\includegraphics[width = 0.195\linewidth ,height=0.066\textheight]{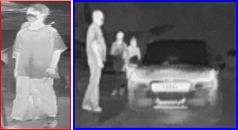} & 
			\hspace{-0.46cm}
			\includegraphics[width = 0.195\linewidth ,height=0.066\textheight]{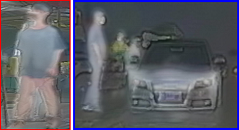}  & 
			\hspace{-0.46cm}
			\includegraphics[width = 0.195\linewidth ,height=0.066\textheight]{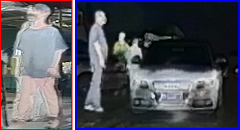}& 
			\hspace{-0.46cm}
			\includegraphics[width = 0.195\linewidth ,height=0.066\textheight]{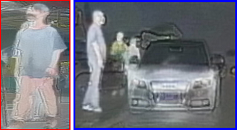} \\ 
   			\footnotesize IR
			& \hspace{-0.46cm} \footnotesize FGAN~\cite{FGAN}
			& \hspace{-0.46cm} \footnotesize DIDFuse~\cite{DIDFuse_2020}
			& \hspace{-0.46cm} \footnotesize PMGI~\cite{PMGI} \\

			\includegraphics[width = 0.195\linewidth ,height=0.066\textheight]{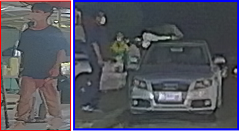} &
			\hspace{-0.46cm}
			\includegraphics[width = 0.195\linewidth ,height=0.066\textheight]{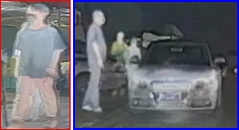}  & 
			\hspace{-0.46cm}
			\includegraphics[width = 0.195\linewidth ,height=0.066\textheight]{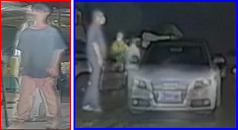}& 
			\hspace{-0.46cm}
			\includegraphics[width = 0.195\linewidth ,height=0.066\textheight]{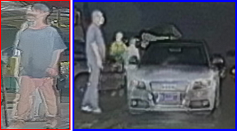} \\ 
                \footnotesize VIS
			& \hspace{-0.46cm} \footnotesize MFEIF~\cite{MFEIF}
			&\hspace{-0.46cm}  \footnotesize RFN~\cite{RFN}
                & \hspace{-0.46cm} \footnotesize U2Fusion~\cite{U2Fusion} \\
                
			\includegraphics[width = 0.195\linewidth ,height=0.066\textheight]{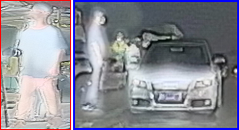} & \hspace{-0.46cm}
			\includegraphics[width = 0.195\linewidth ,height=0.066\textheight]{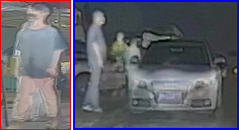} & \hspace{-0.46cm}
                \includegraphics[width = 0.195\linewidth ,height=0.066\textheight]{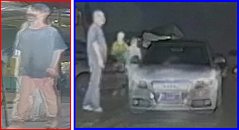} & \hspace{-0.46cm}
			\includegraphics[width = 0.195\linewidth ,height=0.066\textheight]{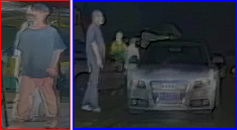}  \\						
			\footnotesize DDcGAN~\cite{DDcGAN}
			& \hspace{-0.46cm} \footnotesize GANMcC~\cite{GANMcC}
                & \hspace{-0.46cm} \footnotesize UMF~\cite{UMF}
			& \hspace{-0.46cm} \footnotesize IRFS (Ours) \\
		\end{tabular}}
	\end{center}
         \vspace{-4mm}
	\caption{Qualitative fusion results of our IRFS versus state-of-the-art infrared-visible image fusion methods on M$^3$FD dataset.}
	\label{fig:fusion-m3fd}
	\vspace{-4mm}
\end{figure*}

As reported in Table~\ref{tab:fusion}, numerically, our IRFS outperforms existing infrared-visible image fusion methods by large margins on VIF and CC metrics.
Indicated that, under the reverse push of the SOD task, the fused image generated by our IRFS can better preserve the information transferred from the source images.
Although the quantitative results on the TNO dataset fail to rank first on the MI metric, they are still favorable considering the direct generalization evaluation of our IRFS.
To give an intuitive evaluation of our IRFS, we exhibit fused results of all discussed methods on three fusion datasets (i.e., TNO, RoadScene, and M$^3$FD), which are shown in Figure~\ref{fig:fusion-road}, Figure~\ref{fig:fusion-tno}, and Figure~\ref{fig:fusion-m3fd}, respectively.
Observing the local enlarged regions of each image, we can find that the results of FGAN~\cite{FGAN}, RFN~\cite{RFN}, DDcGAN~\cite{DDcGAN}, and GANMcC~\cite{GANMcC} all have a serious blurry object and over-smoothed background.
DIDFuse~\cite{DIDFuse_2020} and U2F~\cite{U2Fusion} bring some observable noise and artifacts in their fused results.
In contrast, the fused image generated by IRFS shows a more salient and sharper object that is beneficial to the subsequent SOD task, and it also exhibits cleaner background without extra interference. 

\subsection{Generalization Analysis on SOD}

\begin{figure*}[t]
	\begin{center}
		\resizebox{1.0\linewidth}{!}{
		\begin{tabular}{ccccccccccc}	
			
			\includegraphics[width = 0.092\linewidth]{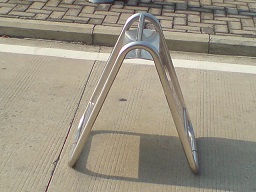}     & \hspace{-0.38cm}
			\includegraphics[width = 0.092\linewidth]{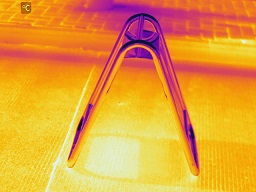}       & \hspace{-0.38cm}
			\includegraphics[width = 0.092\linewidth]{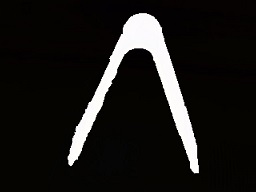}      & \hspace{-0.38cm}
			\includegraphics[width = 0.092\linewidth]{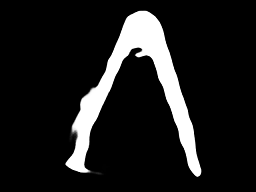} & \hspace{-0.38cm}
			\includegraphics[width = 0.092\linewidth]{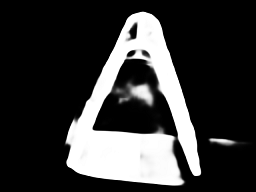}    & \hspace{-0.38cm}
			\includegraphics[width = 0.092\linewidth]{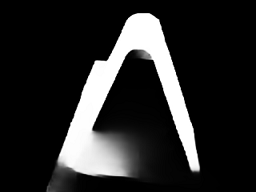}  & \hspace{-0.38cm}
			\includegraphics[width = 0.092\linewidth]{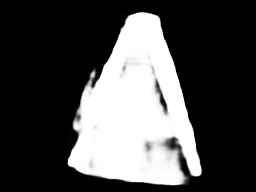}   & \hspace{-0.38cm}
			\includegraphics[width = 0.092\linewidth]{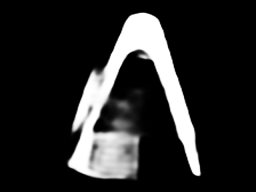} & \hspace{-0.38cm}
			\includegraphics[width = 0.092\linewidth]{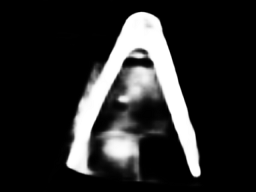}     & \hspace{-0.38cm}
			\includegraphics[width = 0.092\linewidth]{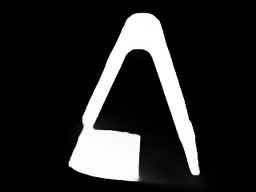} \\
			
			\includegraphics[width = 0.092\linewidth]{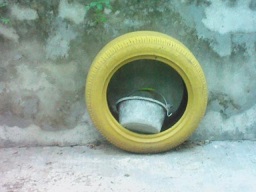}     & \hspace{-0.38cm}
			\includegraphics[width = 0.092\linewidth]{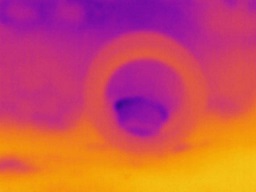}       & \hspace{-0.38cm}
			\includegraphics[width = 0.092\linewidth]{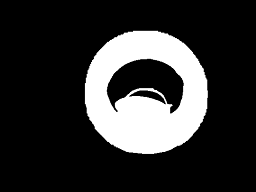}      & \hspace{-0.38cm}
			\includegraphics[width = 0.092\linewidth]{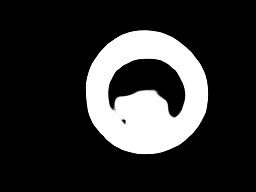} & \hspace{-0.38cm}
			\includegraphics[width = 0.092\linewidth]{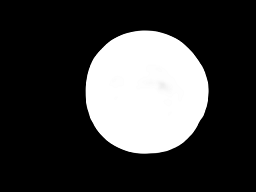}    & \hspace{-0.38cm}
			\includegraphics[width = 0.092\linewidth]{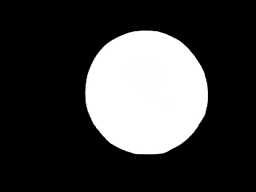}  & \hspace{-0.38cm}
			\includegraphics[width = 0.092\linewidth]{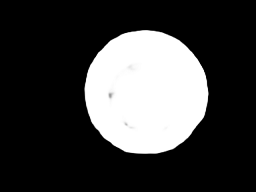}   & \hspace{-0.38cm}
			\includegraphics[width = 0.092\linewidth]{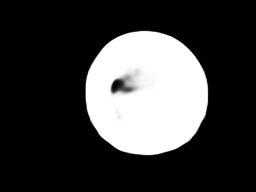} & \hspace{-0.38cm}
			\includegraphics[width = 0.092\linewidth]{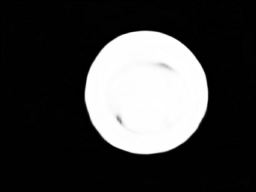}     & \hspace{-0.38cm}
			\includegraphics[width = 0.092\linewidth]{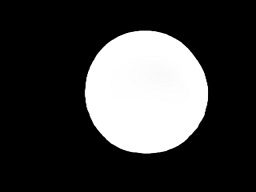} \\
			
			\includegraphics[width = 0.092\linewidth]{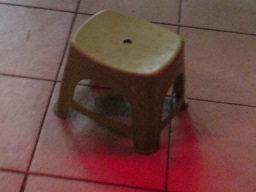}     & \hspace{-0.38cm}
			\includegraphics[width = 0.092\linewidth]{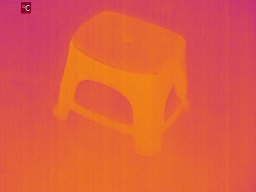}       & \hspace{-0.38cm}
			\includegraphics[width = 0.092\linewidth]{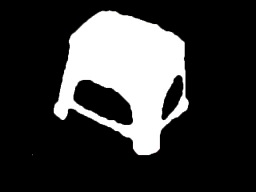}      & \hspace{-0.38cm}
			\includegraphics[width = 0.092\linewidth]{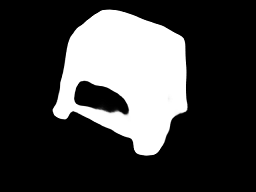} & \hspace{-0.38cm}
			\includegraphics[width = 0.092\linewidth]{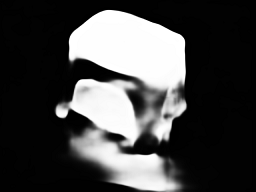}    & \hspace{-0.38cm}
			\includegraphics[width = 0.092\linewidth]{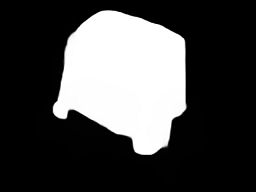}  & \hspace{-0.38cm}
			\includegraphics[width = 0.092\linewidth]{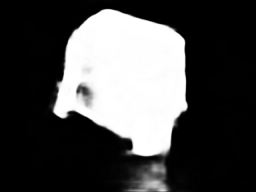}   & \hspace{-0.38cm}
			\includegraphics[width = 0.092\linewidth]{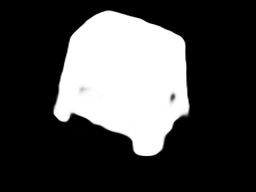} & \hspace{-0.38cm}
			\includegraphics[width = 0.092\linewidth]{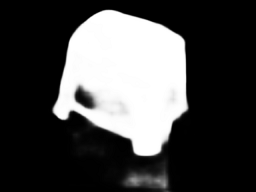}     & \hspace{-0.38cm}
			\includegraphics[width = 0.092\linewidth]{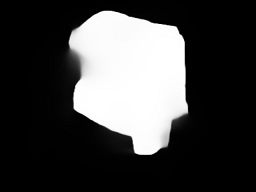} \\
			
			\small Visible
			& \hspace{-0.36cm} \small Infrared
			& \hspace{-0.36cm} \small GT
			& \hspace{-0.36cm} \small IRFS
			& \hspace{-0.36cm} \small MIDD
			& \hspace{-0.36cm} \small CSRN
			& \hspace{-0.36cm} \small MMNet
			& \hspace{-0.36cm} \small ECFFN
			& \hspace{-0.36cm} \small MIA
			& \hspace{-0.36cm} \small OSRN \\
		\end{tabular}}
	\end{center}
	\vspace{-3mm}
	\caption{Visual comparisons versus state-of-the-art SOD methods in some challenging cases: thermal crossover and low contrast.}
	\label{fig:saliency}
	\vspace{-4mm}
\end{figure*}

We conduct the generalization analysis on the SOD task through the comparison with $10$ state-of-the-art thermal infrared and visible SOD methods, including MTMR~\cite{MTMR_18}, M3S-NIR~\cite{M3S-NIR_19}, SGDL~\cite{SGDL_20}, ADF~\cite{ADF_20}, MIDD~\cite{MIDD_21}, CSRN~\cite{CSRN_21}, MMNet~\cite{MMNet_21}, OSRNet~\cite{OSRNet_22}, ECFFN~\cite{ECFFNet_22}, and MIA~\cite{MIA_22}.
Table~\ref{tab:sod} objectively shows the quantitative results measured by four common metrics.
It can be seen that the proposed IRFS ranks either first or second on these three datasets.
For instance, compared with the second best method OSRNet on the VT5000 dataset, the gain reaches 2.1\% for mean $F_{\beta}$, 1.6\% for mean $E_{\xi}$, and 13.16\% for MAE score. In terms of mean $S_{\alpha}$ metric, our result is only 0.11\% less than OSRNet.
To better reflect the superiority of our IRFS, as shown in Figure~\ref{fig:saliency}, we visualize the saliency maps predicted by all the aforementioned methods.
It is clearly observed that our IRFS obtains more accurate saliency maps with less false detection compared with OSRNet, the latest study.
Taking the second image as an example, Although other SOD methods also localize the \textit{ring} of the image and predict an accurate circular outline, they have difficulty identifying the small object inside the \textit{ring}.
In contrast, our IRFS not only highlights the structural integrity of salient objects but also guarantees internal consistency within the object contours.
It is indicated that our IRFS is more robust to some challenging scenarios, such as thermal crossover, small objects, and low contrast.

\begin{table*}[tb!]
	\centering
	\caption{Quantitative SOD results. $\uparrow$/$\downarrow$ for a metric denotes that a larger/smaller value is better. The best results are bolded and the second-best results are highlighted in \underline{underline}.} 
	\label{tab:sod}
	\renewcommand{\arraystretch}{0.8}
	\renewcommand{\tabcolsep}{1.4mm}
	\resizebox{1.0\linewidth}{!}{
	\begin{tabular}{lrccccccccccc}
		\hline
		\toprule
		&\multirow{3}{*}{}\centering  & \textbf{\small MTMR}   & \textbf{\small M3S-NIR} & \textbf{\small SGDL} & \textbf{\small ADF} & \textbf{\small MIDD} & \textbf{\small CSRN} & \textbf{\small MMNet} & \textbf{\small OSRNet} &  \textbf{\small ECFFN} & \textbf{\small MIA} & \textbf{\small IRFS}\\
		\specialrule{0em}{1pt}{1pt}
		\hline
		\specialrule{0em}{1pt}{1pt}
		\multirow{3}{*}{\textit{}}
		\multirow{5}{*}{\textit{VT821}}
		& $S_{\alpha}\uparrow$ & 0.725 & 0.723 & 0.765 & 0.810  & 0.871 & \textbf{0.879}  &0.873 & 0.875 & \underline{0.877} & 0.844 &  \textbf{0.879}\\
		& $F_{\beta}\uparrow$  & 0.662 & 0.734 & 0.730 & 0.716 & 0.804 & \underline{0.830} & 0.794 & 0.813 &0.810 & 0.740 & \textbf{0.833} \\
		& $E_{\xi}\uparrow$ & 0.815 & 0.859 & 0.847 & 0.842 & 0.895 & \underline{0.908} & 0.892 & 0.896 & 0.902 & 0.850 &\textbf{0.917}\\
		& $\mathcal{M}\downarrow$ & 0.108 & 0.140 & 0.085 & 0.077 & 0.045 & 0.038 & 0.040 & 0.043 & \underline{0.034} & 0.070 &\textbf{0.029}\\
		\midrule
		\multirow{5}{*}{\textit{VT1000}}
		& $S_{\alpha}\uparrow$ & 0.706 & 0.726 & 0.787 & 0.910 & 0.907 & 0.918 & 0.914 & \textbf{0.926} & 0.923 & \underline{0.924} & \underline{0.924} \\
		& $F_{\beta}\uparrow$  & 0.715 & 0.717 & 0.764 & 0.847 & 0.871 & 0.877 & 0.861 & \underline{0.892} & 0.876 & 0.868 & \textbf{0.901} \\
		& $E_{\xi}\uparrow$ & 0.836 & 0.827 & 0.856 & 0.921 & 0.928 & 0.925 & 0.923 & \underline{0.935} & 0.930 & 0.926 &\textbf{0.943} \\
		& $\mathcal{M}\downarrow$ & 0.119 & 0.145 & 0.090 & 0.034 & 0.029 & 0.024 & 0.027 & 0.022 & \underline{0.021} & 0.025 & \textbf{0.019} \\
		\midrule
		\multirow{5}{*}{\textit{VT5000}}
		& $S_{\alpha}\uparrow$ & 0.680 & 0.652 & 0.750 & 0.863 & 0.856 & 0.868 & 0.862 & 0.875 & 0.874 & \textbf{0.878} & \underline{0.877} \\
		& $F_{\beta}\uparrow$  & 0.595 & 0.575 & 0.672 & 0.778 & 0.789 & 0.810 & 0.780 & \underline{0.823} & 0.806 & 0.793 & \textbf{0.835} \\
		& $E_{\xi}\uparrow$   & 0.795 & 0.780 & 0.824 & 0.891 & 0.891 & 0.905 &  0.887& \underline{0.908} & 0.906 & 0.893 &\textbf{0.922} \\
		& $\mathcal{M}\downarrow$ & 0.114 & 0.168 & 0.089 & 0.048 &  0.046 & 0.042 & 0.043 & 0.040 & \underline{0.038} & 0.040 &\textbf{0.034} \\
		\bottomrule
		\hline
	\end{tabular}}
\end{table*}

\subsection{Analysis of Model Efficiency}
For a multi-task framework, the efficiency of the model is crucial due to the requirement of real-time application. We analyze the model size (Mb) and inference time (s) of all comparison methods, including several state-of-the-art (SOTA) fusion methods and multimodal SOD based on thermal infrared and visible images.
Due to the multi-task settings of our IRFS framework, we ensure a fair comparison by evaluating the fusion module, FSFNet, against SOTA image fusion methods, and concurrently evaluating the SOD module, FGC$^2$Net, against SOTA thermal infrared-visible SOD methods.
Noted that each model is conducted on a single NVIDIA 1080Ti GPU with input being resized to $640 \times 480$.
According to Table~\ref{tab:efficiency}, our FSFNet ranks second in terms of model size, exhibiting a relatively low parameter count of $0.06$Mb. Our FGC$^2$Net has $39.71$Mb parameters and ranks third. Noted that MTMR~\cite{MTMR_18}, SGDL~\cite{SGDL_20}, and M3S-NIR~\cite{M3S-NIR_19} are traditional SOD methods, we fail to provide their model sizes.
Combined with Table~\ref{tab:sod}, CSRN~\cite{CSRN_21} has the smallest model size, but it does come at the cost of performance.
Although not the most compact model, our FSFNet and FGC$^2$Net have the fastest inference speed. The core reason, the one is that a dual attention-guided feature screening module of FSFNet replaces frequently-used dense network, and the other is that a siamese encoder based on the lightweight ResNet-34 model is used as the backbone of FGC$^2$Net. Overall, the proposed IRFS framework reaches a better balance between performance and efficiency.

\subsection{Discussion for Weight $\tau$ of Fusion Loss}\label{subsec:tau}
To examine the influence of different trad-off weights $\tau$ of fusion loss $\mathcal{L}_{\texttt{fusion}}$ on image fusion and SOD tasks, we set $\tau$ to $0.1$, $0.5$, $1.0$, $5.0$, and $10$ to train our IRFS framework on the VT5000 dataset, respectively. Table~\ref{weight} reports the quantitative results of both tasks. 
The first two rows show fused results evaluated on the CC and VIF metrics. The last two rows show SOD accuracy measured by the $S_{\alpha}$ and $F_{\beta}$ metrics. According to the analysis in Table~\ref{weight}, we can find that, when $\tau=1.0$, fusion results exhibit a noticeable superiority compared to those achieved with other alternative $\tau$ values.
For the evaluation of the SOD task, although the precision of $F_{\beta}$ reaches its peak when $\tau=0.5$, it only improves by $0.004$ compared to when $\tau=1.0$. 
Therefore, we empirically set $\tau$ to $1.0$ in Eq.~(\ref{equ:joint_loss}).

\begin{table*}[t]
        \caption{Efficiency analysis of our IRFS framework. To be fair, the efficiency evaluations against state-of-the-art thermal infrared-visible fusion methods and SOD models are implemented, respectively. $*$ indicates that the current model is part of our IRFS framework.}
	\label{tab:efficiency}
        \vspace{-0.3cm}
	\begin{center}
        \renewcommand{\arraystretch}{0.9}
	\renewcommand{\tabcolsep}{2.0mm}
        \resizebox{1.0\linewidth}{!}{
        \begin{tabular}{llcccccccccc}
        \hline
            \toprule
            
            &  & \textbf{FGAN} & \textbf{DIDFuse} & \textbf{PMGI} & \textbf{U2F} & \textbf{MFEIF} & \textbf{RFN} & \textbf{DDcGAN} & \textbf{GANMcC} & \textbf{UMF} & \textbf{FSFNet$^*$} \\
            \midrule
            \specialrule{0em}{1pt}{1pt}
            \multirow{2}{*}{\textbf{\textit{Fusion}}} & Size (Mb)  & $0.074$ & $0.261$ & $0.042$ & $0.659$ & $0.370$ & $10.94$ & $1.097$ & $1.864$ & $0.80$ & $0.060$\\
            \specialrule{0em}{1pt}{1pt}
            & Time (\textit{s}) & $0.122$ & $0.053$ & $0.058$ & $0.057$ & $0.098$ & $0.135$ & $0.712$ & $0.338$ & $0.057$ & $0.025$ \\
            \midrule
            &  & \textbf{MTMR} & \textbf{SGDL} & \textbf{M3S-NIR} &\textbf{ADF} & \textbf{MIDD} & \textbf{CSRN} & \textbf{MMNet} & \textbf{OSRNet} & \textbf{MIA} & \textbf{FGC$^2$Net$^*$} \\
            \midrule
            \specialrule{0em}{1pt}{1pt}
            \multirow{2}{*}{\textbf{\textit{SOD}}} & Size (Mb)  & $--$ & $--$ & $--$ & $83.13$ & $52.43$ & $1.010$ & $64.36$ & $15.64$& $285.7$ & $39.71$\\
            \specialrule{0em}{1pt}{1pt}
            & Time (\textit{s}) & $2.515$ & $2.230$ & $1.500$ & $0.223$ & $0.126$ & $0.050$ & $0.079$ & $0.040$ & $0.045$ & $0.034$ \\
            \bottomrule
            \hline
		\end{tabular}}
		\vspace{-2mm}
		
	\end{center}
	\vspace{-4mm}
\end{table*}

\begin{table}[h]
	\vspace{-4mm}
	\caption{Discussion for the weight $\tau$ of the fusion loss $\mathcal{L}_{\texttt{fusion}}$. Two fusion metrics (i.e., CC and VIF) and two SOD metrics (i.e., $S_{\alpha}$ and $F_{\beta}$) are used to evaluate fusion results and saliency maps.}\label{weight}
	\centering
	\renewcommand{\arraystretch}{1.1}
	\setlength{\tabcolsep}{1.8mm}{
		\resizebox{1.0\linewidth}{!}{
		\begin{tabular}{lcccccc}
            \hline
		\toprule
			Metric & $\tau=0.1$ & $\tau=0.5$ & $\tau=1.0$ & $\tau=2.0$ & $\tau=5.0$ & $\tau=10$\\
		\hline
            \specialrule{0em}{1pt}{1pt}
            $\text{CC}\uparrow$     & 1.431 & 1.376 & \textbf{1.450} & 1.417 & 1.371 & 1.369 \\
            $\text{VIF}\uparrow$    & 0.977 & 1.010 & \textbf{1.684} & 0.974 & 0.892 & 0.940 \\
            \midrule
            $S_{\alpha}\uparrow$    & 0.871 & 0.876 & \textbf{0.877} & 0.873 & 0.875 & 0.872 \\
		$F_{\beta}\uparrow$     & 0.837 & \textbf{0.839} & 0.835 & 0.834 & 0.830 & 0.821 \\
		\bottomrule
            \hline
	\end{tabular}}}
	\vspace{-2mm}
\end{table}

\begin{table}[t]
	\centering
        \vspace{-2mm}
	\caption{Investigation of the interactive loop learning strategy from the fusion perspective on \textbf{RoadScene} dataset.}
	\label{tab:inter_fusion}
	\renewcommand{\arraystretch}{1.0}
	\setlength{\tabcolsep}{3.5mm}{
		\resizebox{1.0\linewidth}{!}{
		\begin{tabular}{r|c|c|c|c}
			\toprule
			\multirow{2}{*}{Metric} & \multirow{2}{*}{One-stage}  & \multicolumn{3}{c}{Interaction} \\ 
			\cline{3-5} 
			&   & 1-$st$ &  5-$th$ & 9-$th$ \\ 
			\hline
			\multicolumn{1}{l|}{$\text{MI}\uparrow$}       & 2.149 & 2.178 & 2.193 & \textbf{2.301}\\ 
			\multicolumn{1}{l|}{$\text{VIF}\uparrow$}      & 1.047 & 1.141 & 1.175 & \textbf{1.223} \\ 
			\multicolumn{1}{l|}{$\text{CC}\uparrow$}       & 1.136 & 1.158 & 1.162 & 
			\textbf{1.164} \\
			\bottomrule
	\end{tabular}}}
	\vspace{-2mm}
\end{table}

\subsection{Ablation Studies}

\subsubsection{Interactive training~\textit{vs.}~One-stage training}
\begin{figure}[h]
	\centering
\resizebox{1.0\linewidth}{!}{
	\begin{tabular}{c}
		\includegraphics[width = 0.99\linewidth]{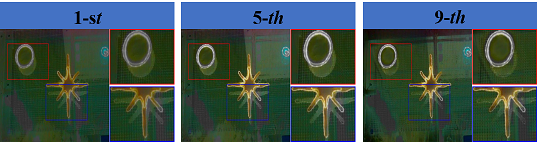} \\
		(a) Fusion results of different interactive loops \\
		\includegraphics[width = 0.99\linewidth]{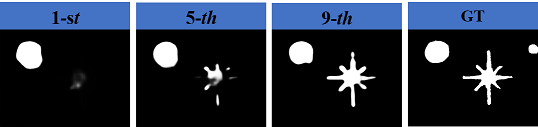}\\
		(b) SOD results of different interactive loops \\
	\end{tabular}}
	\caption{Qualitative analysis of the interactive loop learning strategy from fusion and SOD perspectives on the VT5000 dataset, respectively.}%
	\vspace{-4mm}
	\label{fig:joint_fus_sod}
\end{figure}

To illustrate the effectiveness of the interactive loop learning strategy, we implement the comparisons between our interactive training strategy and one-stage training strategy from two perspectives of fusion and SOD, respectively.
Since our IRFS interactively implements FSFNet and FGC$^2$Net for $10$ loops, denoted from $0$ to $9$, we present the quantitative results of fusion and SOD at three intervals, i.e., 1-$st$, 5-$th$, and 9-$th$, as shown in Table~\ref{tab:inter_fusion} and Table~\ref{tab:inter_sod}.
We can find that the proposed interactive loop learning strategy is indeed more conducive to the mutual promotion of infrared-visible image fusion and SOD tasks. And, with the increase of loops, the respective performance of both fusion and SOD keeps improving.
It is suggested that there is a collaborative relationship between infrared-visible image fusion and SOD tasks, and both of them are expected to be deployed in a single framework to reach the purpose of locating salient objects and returning a high-quality fused image.
To explicitly evaluate the effectiveness of the proposed interactive loop learning strategy, we exhibit fusion and sod results of different interactive loops.
In our implementation, the interactive learning process lasts for $10$ epoch from 0 to the 9-\textit{th} epoch.
We show the fusion and SOD results of the 1-\textit{st}, 5-\textit{th}, and 9-\textit{th} epoch in Figure~\ref{fig:joint_fus_sod}, which is corresponding to the Table~\ref{tab:inter_fusion} and Table~\ref{tab:inter_sod}.
Observing these locally enlarged regions of fusion results in Figure~\ref{fig:joint_fus_sod}(a), as the interactive loop training deepens, the salient objects are highlighted and the interference of background is weakened progressively, which is conducive to localizing these salient objects.
This conclusion is supported by these SOD results in Figure~\ref{fig:joint_fus_sod}(b).

\begin{table}[t]
	\centering
 \vspace{-4mm}
	\caption{Investigation of the interactive loop learning strategy from the SOD perspective on \textbf{VT5000} dataset.}
	\label{tab:inter_sod}
	\renewcommand{\arraystretch}{1.0}
	\setlength{\tabcolsep}{3.5mm}{
		\resizebox{1.0\linewidth}{!}{
		\begin{tabular}{c|c|c|c|c}
			\toprule
			\multirow{2}{*}{Metric} & \multirow{2}{*}{One-stage}  & \multicolumn{3}{c}{Interaction} \\ 
			\cline{3-5} 
			&   & 1-$st$ &  5-$th$ & 9-$th$ \\ 
			\hline
			\multicolumn{1}{l|}{$S_{\alpha}\uparrow$}     & 0.863 & 0.869 & 0.874 & \textbf{0.877} \\ 
			\multicolumn{1}{l|}{$F_{\beta}\uparrow$}      & 0.803 & 0.824 & 0.830 & \textbf{0.835} \\ 
			\multicolumn{1}{l|}{$E_{\xi}\uparrow$}        & 0.907 & 0.913 & 0.919 & \textbf{0.922} \\
			\multicolumn{1}{l|}{$\mathcal{M}\downarrow$}  & 0.038 & 0.037 & 0.035 & \textbf{0.034} \\
			\bottomrule
	\end{tabular}}}
	\vspace{-2mm}
\end{table}

\begin{table}[h]
	\caption{Effectiveness analysis of the fused image for the SOD task on \textbf{VT5000} dataset.}
	\label{tab:ab-fusion}
	\centering
	\renewcommand{\arraystretch}{1.15}
	\setlength{\tabcolsep}{4.0mm}{
		\resizebox{1.0\linewidth}{!}{
		\begin{tabular}{l|cccc}
			\toprule
			Metric & $I_{r}$ & $I_{t}$ & $\left(I_{r}+I_{t}\right)/2$ & $I_{f}$ \\
			\hline
			\specialrule{0em}{1pt}{1pt}
			$S_{\alpha}\uparrow$    & 0.870 & 0.869 & 0.872 & \textbf{0.877} \\
			$F_{\beta}\uparrow$     & 0.799 & 0.808 & 0.830 & \textbf{0.835} \\
			$E_{\xi}\uparrow$       & 0.903 & 0.802 & 0.920 & \textbf{0.922} \\
			$\mathcal{M}\downarrow$ & 0.038 & 0.036 & 0.036 & \textbf{0.034} \\
			\bottomrule
	\end{tabular}}}
	\vspace{-2mm}
\end{table}

\begin{table}[t]
	\centering
	\vspace{1mm}
	\caption{Ablation studies of key components of the proposed FGSE module.}\label{tab:ab_fgse}
	\footnotesize
	\renewcommand{\arraystretch}{1.0}
	\renewcommand{\tabcolsep}{1.6mm}
	\resizebox{1.0\linewidth}{!}{
	\begin{tabular}{l|cccccccc}
		\toprule
		\multirow{2}{*}{\#} 
		& \multirow{2}{*}{SEM} & \multirow{2}{*}{C$^2$FTL} & \multirow{2}{*}{LFS} & &\multicolumn{4}{c}{\bf VT5000}\\
		&
		&
		&
		&
		& $S_{\alpha}\uparrow$
		& $F_{\beta}\uparrow$
		& $E_{\xi}\uparrow$ 
		& $\mathcal{M}\downarrow$ \\
		\midrule
		0 & \checkmark &  &  && 0.868 & 0.819 & 0.915 & 0.036 \\
		1 & \checkmark &  & \checkmark && 0.869 & 0.826 & 0.914 & 0.037 \\
		2 & \checkmark & \checkmark &  && 0.872 & 0.825 & 0.918 & 0.035 \\
		\midrule
		3 & \checkmark & \checkmark & \checkmark && \textbf{0.877} & \textbf{0.835} & \textbf{0.922} & \textbf{0.034} \\
		\bottomrule
	\end{tabular}}
	\vspace{-2mm}
\end{table}

\subsubsection{Effectiveness of pixel-level fusion for SOD}
\begin{figure}[t]
	\begin{center}
			\begin{tabular}{ccc}	
				\includegraphics[width = 0.32\linewidth]{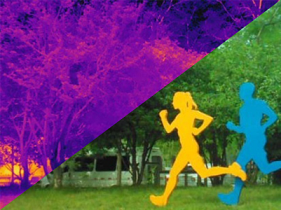} &
                \hspace{-0.46cm}
				\includegraphics[width = 0.32\linewidth]{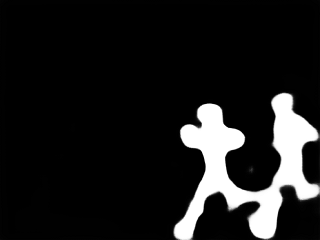} & 
				\hspace{-0.46cm}
				\includegraphics[width = 0.32\linewidth]{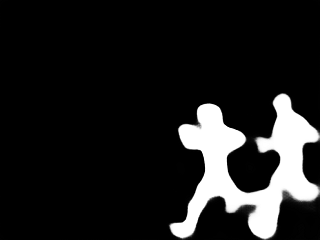} \\
                (a) IR/VIS	&
                \hspace{-0.46cm} (b) $I_{r}$ & 
                \hspace{-0.46cm} (c) $I_{t}$ \\

				\includegraphics[width = 0.32\linewidth]{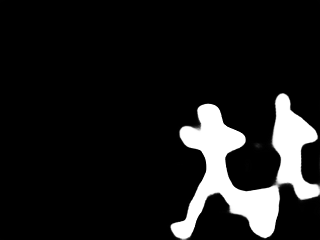} & 
				\hspace{-0.46cm}
				\includegraphics[width = 0.32\linewidth]{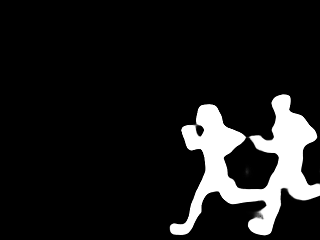} &
                    \hspace{-0.46cm}
                    \includegraphics[width = 0.32\linewidth]{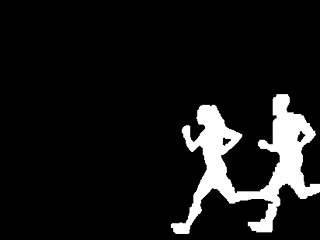} \\	
                (d)	$\left(I_{r}+I_{t}\right)/2$ &
                \hspace{-0.46cm} (e) $I_{f}$ & 
                \hspace{-0.46cm} (f) GT \\
				
				
		\end{tabular}
	\end{center}
	\vspace{-2mm}
	\caption{Qualitative analysis of the effectiveness of the fused image for the SOD task on VT5000 dataset.}
	\label{fig:ab-fusion}
	\vspace{-2mm}
\end{figure}

One core motivation is to explore the relationship between pixel-level fusion and SOD tasks.
To demonstrate whether the fused image has a boosting effect on SOD performance, we utilize source images $I_{r}$, $I_{t}$, and their weighted-fusion variant $\left(I_{r}+I_{t}\right)/2$ instead of the fused image $I_{f}$ generated by our FSFNet in IRFS framework, respectively, and treat them as the third input of the FGC$^2$Net to perform the SOD task.
As shown in Table~\ref{tab:ab-fusion}, when the rough fused image $\left(I_{r}+I_{t}\right)/2$ from source RGB and thermal infrared images is used as a guidance of the SOD task, the quantitative values in four metrics including $S_{\alpha}$, $F_{\beta}$, $E_{\xi}$, and MAE outperform those of the implementations regarding RGB or thermal infrared images as the guidance of FGC$^2$Net.
And further, when our fused image $I_{f}$ generated by FSFNet is adopted to guide the subsequent FGC$^2$Net, these quantitative results achieve more desirable improvements.
Therefore, we believe that image fusion is beneficial to facilitating SOD networks to predict more accurate saliency maps.
Figure~\ref{fig:ab-fusion} shows the qualitative results corresponding to Table~\ref{tab:ab-fusion} to explicitly investigate the effectiveness of the pixel-level fusion for the subsequent SOD task.
It is easy to observe that using our fusion image as a third modality to guide the SOD task can predict more precise saliency maps with coherent edges and complete objects.
Coupled with the quantitative results, we can argue that the pixel-level fusion results are beneficial to facilitating SOD.

\begin{figure}[t]
	\begin{center}
			\begin{tabular}{ccc}	
				\includegraphics[width = 0.32\linewidth]{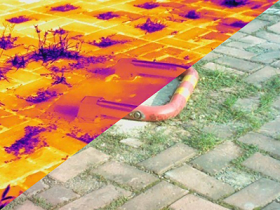} &
                \hspace{-0.46cm}
				\includegraphics[width = 0.32\linewidth]{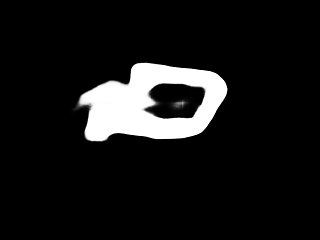} & 
				\hspace{-0.46cm}
				\includegraphics[width = 0.32\linewidth]{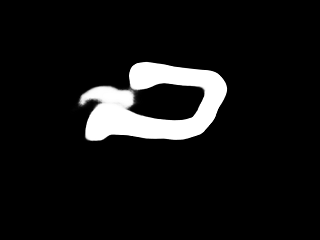} \\
                \small (a) IR/VIS	&
                \hspace{-0.46cm} \small (b) w/o C$^2$FTL, LFS & 
                \hspace{-0.46cm} \small (c) w/o C$^2$FTL \\

				\includegraphics[width = 0.32\linewidth]{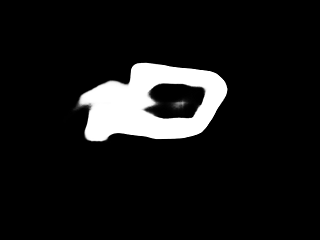} & 
				\hspace{-0.46cm}
				\includegraphics[width = 0.32\linewidth]{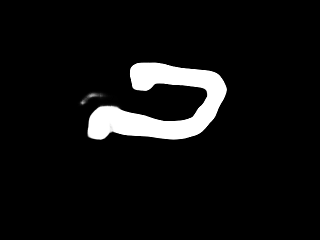} &
                    \hspace{-0.46cm}
                    \includegraphics[width = 0.32\linewidth]{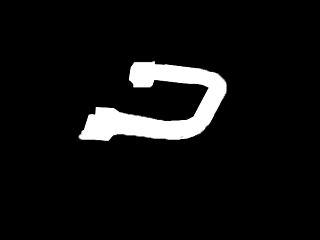} \\	
                (d)	w/o LFS &
                \hspace{-0.46cm} (e) IRFS & 
                \hspace{-0.46cm} (f) GT \\
		\end{tabular}
	\end{center}
	\vspace{-2mm}
	\caption{Ablation studies of key components of the proposed FGSE module on VT5000 dataset, qualitatively.}
	\label{fig:ab-fgse}
	\vspace{-2mm}
\end{figure}

\subsubsection{Effectiveness of FGSE}
We mainly focus on the investigations of key components of the FGSE module (shown in Figure~\ref{fig:ab-fgse}).
To verify the effectiveness of C$^2$FTL and LFS, we successively remove them from the FGSE module, denoted as \#2 and \#1 in Table~\ref{tab:ab_fgse}.
It can be seen that using the C$^2$FTL brings an improvement of 12.1\% in the MAE score and using the LFS brings an improvement of 1.9\% in mean $F_{\beta}$ metric on the VT5000 dataset.
Assuming that both C$^2$FTL and LFS are absent, denoted as \#0, the performance drops by more than 2.5\% in $F_{\beta}$ and 9.0\% in MAE.
The results comprehensively reveal the contribution of our FGSE module for the enhancement of the saliency-related features and the suppression of the interference information on the SOD task.
Figure~\ref{fig:ab-fgse} shows the qualitative results corresponding to Table~\ref{tab:ab_fgse} to explicitly examine the effectiveness of key components of the FGSE module in the SOD task.
We can observe that the removal of the C$^2$FTL or LSF leads to the adhesion of salient objects in predicted saliency maps.
By contrast, our IRFS predicts more precise saliency maps with sharp object outlines.
The results reveal the effectiveness of the proposed FGSE module.

\section{Conclusion}
This paper first proposed an interactively reinforced paradigm for joint infrared-visible image fusion and saliency object detection tasks, specifically designed for the unmanned system to search and parse objects in the wild. 
In this paradigm, image fusion focuses on highlighting saliency-related features to suit the requirements of the infrared-visible SOD task, while the SOD task propagates semantic loss back to the fusion part and implements supervision to prevent the generated fused images from losing semantic information.
The comprehensive experimental results revealed that infrared-visible image fusion and SOD tasks can maintain a collaborative relationship in a single framework.
The proposed paradigm has two notable strengths. Firstly, the resulting interactive reinforcement between the two tasks leads to improved performance in both infrared-visible image fusion and SOD tasks. Secondly, our paradigm represents a significant contribution to the visual perception of unmanned systems. Nevertheless, there is still room for improvement. Specifically, the paradigm has yet to be adapted to dynamically adjust image fusion to meet the practical requirements of SOD in challenging scenarios such as low-light or adverse weather conditions.
Addressing this limitation is an important area for future research, as it would significantly enhance the utility of the framework in real-world applications.

\section{Acknowledgements}
This work is partially supported by the National Key R\&D Program
of China (2020YFB1313503 and 2022YFA1\-004101), the National
Natural Science Foundation of China (Nos. U22B2052 and 61922019).


\bibliographystyle{cas-model2-names}

\bibliography{cas-refs}

\bio{}
\endbio


\end{document}